\documentclass[10pt,journal]{IEEEtran}

\usepackage{amsmath,amssymb,amsfonts}
\usepackage{graphicx}
\usepackage{subfig}
\usepackage[hyphens]{url}
\usepackage{multirow}
\usepackage{float}
\usepackage{booktabs}
\usepackage{pifont}
\usepackage{todonotes}
\usepackage{adjustbox}
\usepackage{threeparttable}
\usepackage{algorithm}
\usepackage{algorithmic}

\usepackage[hidelinks]{hyperref}
\hypersetup{
pdfauthor={P. Drozdowski, C. Rathgeb, F. Stockhardt, D. Osorio-Roig, C. Busch},
pdfsubject={Biometrics},
pdftitle={Feature Fusion Methods for Indexing and Retrieval of Biometric Data: Application to Face Recognition with Privacy Protection},
pdfkeywords={Biometric Identification, Biometric Template Protection, Computational Workload Reduction, Indexing, Information Fusion, Face Recognition},
bookmarks=true,
bookmarksopen=true,
bookmarksnumbered=true,
pagebackref=false
}

\ifCLASSOPTIONcompsoc
  \usepackage[nocompress]{cite}
\else
  \usepackage{cite}
\fi

\def\eg{\textit{e.g}. } 
\def\ie{\textit{i.e}. } 
\def\cf{\textit{c.f}. } 
\def\etc{\textit{etc}. } 

\def\wrt{w.r.t. } \def\etal{\textit{et al}. }

\newcommand{\set}[1]{\left\lbrace #1 \right\rbrace}

\newcommand{\cmark}{\ding{51}}%

\begin{document}
\title{Feature Fusion Methods for Indexing and Retrieval of Biometric Data: Application to Face Recognition with Privacy Protection}

\author{\IEEEauthorblockN{Pawel~Drozdowski,
        Fabian~Stockhardt,
        Christian~Rathgeb,
        Dail{\'e} Osorio-Roig,
        Christoph~Busch} \\ \vspace{0.1cm}
    \IEEEauthorblockA{da/sec -- Biometrics and Internet Security Research Group, Hochschule Darmstadt, Germany
    \\\texttt{\{name.lastname\}@h-da.de}
} \vspace{-0.5cm}
}

\IEEEtitleabstractindextext{
\begin{abstract}
Computationally efficient, accurate, and privacy-preserving data storage and retrieval are among the key challenges faced by practical deployments of biometric identification systems worldwide. 

In this work, a method of protected indexing of biometric data is presented. By utilising feature-level fusion of intelligently paired templates, a multi-stage search structure is created. During retrieval, the list of potential candidate identities is successively pre-filtered, thereby reducing the number of template comparisons necessary for a biometric identification transaction. Protection of the biometric probe templates, as well as the stored reference templates and the created index is carried out using homomorphic encryption. 

The proposed method is extensively evaluated in closed-set and open-set identification scenarios on publicly available databases using two state-of-the-art open-source face recognition systems. With respect to a typical baseline algorithm utilising an exhaustive search-based retrieval algorithm, the proposed method enables a reduction of the computational workload associated with a biometric identification transaction by 90\%, while simultaneously suffering no degradation of the biometric performance. Furthermore, by facilitating a seamless integration of template protection with open-source homomorphic encryption libraries, the proposed method guarantees unlinkability, irreversibility, and renewability of the protected biometric data.
\end{abstract}

}

\maketitle

\IEEEdisplaynontitleabstractindextext
\IEEEpeerreviewmaketitle

\section{Introduction}
\label{sec:introduction}
Personal, commercial, and governmental identity management systems increasingly rely on biometric technologies, which enable reliable recognition of individuals based on highly distinctive characteristics of human beings, \eg face or fingerprints. Applications ranging from personal device access \cite{Das-MobileBiometrics-2018}, border control \cite{EULisa-EURODAC-2016,SmartBorders-EU-2018,Thales-IDENT-2021}, forensic investigations and law enforcement \cite{Moses-AFIS-2010,FBI-CODIS-2021,Thales-AFIS-2021}, national ID systems \cite{UIDAI-Aadhaar-2012,Dalwai-Aadhaar-2014}, and voter registration \cite{Bowyer-IrisElection-2015,CEPPS-Congo-2018} benefit from the use of biometrics. The largest systems of this kind enrol hundreds of millions or even beyond a billion enrolled subjects (see \eg \cite{UIDAI-Dashboard}), with the global market value of biometric technologies currently estimated to be tens of billions of dollars \cite{Pascu-BiometricMarketValue-2020}.

As the prevalence, size, and scope of the operational biometric systems increase, the development of technologies which are capable of accurately and efficiently processing biometric data becomes critically important. In the challenging identification and duplicate enrolment check scenarios, where typically an exhaustive search (\ie one-to-many comparison) is needed, solutions which facilitate practical system response times are indispensable. Rather than merely scaling the hardware architecture, which is associated with high monetary costs, algorithmic methods (such as indexing) referred to as biometric workload reduction \cite{Drozdowski-WorkloadSurvey-IET-2019} can be used to speed-up the search queries (and hence reduce the monetary costs). In recent years, strong interest from governmental side in such methods has been manifested through numerous benchmarks and competitions \cite{FRVT-Identification-2020,IREX-2018,Bust-DHSRally-2019}.

In addition to the aforementioned practical requirements pertaining to biometric performance and computational efficiency, preventing misuse (\eg privacy violations) of the stored biometric reference data is essential. Existing privacy regulations, \eg the General Data Protection Regulation (GDPR) \cite{EU-GDPR-2016}, classify biometric data under ``special categories of personal data'', thus entailing significant responsibilities for the data controllers. Traditional encryption methods are unsuitable for protecting biometric data, since biometric characteristics exhibit a natural intra-class variance. If traditional cryptographic techniques are applied to biometric templates, said biometric variance prevents a biometric comparison in the encrypted domain. That is, the use of conventional cryptographic methods would require a decryption of protected biometric data prior to the comparison. In contrast, \emph{biometric template protection} \cite{5396650,Rathgeb-TemplateProtection-EURASIP-2011,Nandakumar-TemplateProtection-IEEE-2015} enables a comparison of biometric data in the encrypted domain and hence a permanent protection of biometric data. Biometric template protection schemes use auxiliary data to obtain pseudonymous identifiers from unprotected biometric data. Biometric comparisons are then performed via pseudonymous identifiers while unprotected biometric reference data is discarded \cite{ISO11-TemplateProtection}. Biometric template protection schemes have hardly been employed in biometric identification systems \cite{Drozdowski-WorkloadSurvey-IET-2019}. One reason for this is that many types of biometric template protection schemes require complex comparison methods which renders them unsuitable for biometric identification (where the workload is dominated by comparison costs). So far, only a handful approaches have combined computational workload reduction strategies with biometric template protection. In the context of face biometrics, those studies have mainly employed cancelable biometrics, \eg \cite{Dong-face-identification-index-2020,Murakami-CancelableIndexing-2019,Sardar-novel-cancelable-face-hashing-2020}. However, most of those systems still report a degradation \wrt biometric performance when benchmarked against unprotected systems. Practically feasible applications of homomorphic encryption in biometric identification systems have likewise been presented \cite{Boddeti-HE-2019,Drozdowski-HomomorphicIdentificationFace-BIOSIG-2019,Kolberg-FaceHE-BTP-BIOSIG-2020,Engelsma-HomomorphicIdentification-2020}; however, while suffering little to none biometric performance degradation, these schemes also have relied on exhaustive search in a biometric identification scenario and have not considered integration of computational workload reduction such as biometric indexing.

\subsection{Contribution and Organisation}
\label{subsec:introduction_organisation}
The main contributions of this article are as follows:

\begin{itemize}
\item A comprehensive overview and literature survey of works pertaining to (and especially those combining) the areas of information fusion, computational efficiency, and template protection in biometric identification systems.
\item A proposal of a multi-stage protected indexing and retrieval system for facial biometric identification based on optimised information fusion and incorporating data privacy-preservation with homomorphic encryption.
\item A thorough theoretical analysis and empirical evaluation of the proposed system on a large dataset with state-of-the-art facial recognition systems. Using ISO/IEC IS 19795-1 \cite{ISO-PerformanceReporting-2021} compliant experimental protocol and metrics, the proposed system is shown to reduce the computational workload of a biometric identification retrieval by approximately 90\%, while simultaneously maintaining the baseline biometric performance. Additionally, the possibility of seamless integration of post-quantum-secure homomorphic encryption means that the data security and privacy objectives specified in ISO/IEC IS 24745 \cite{ISO11-TemplateProtection} are ensured.
\end{itemize}

The remainder of this article is organised as follows: section \ref{sec:background} provides relevant background information and an overview of related works. The proposed system is described in section \ref{sec:proposedsystem}. The experimental setup and the obtained results are presented in sections \ref{sec:experimentalsetup} and \ref{sec:results}, respectively. Section \ref{sec:conclusion} contains concluding remarks and a summary. 

\section{Background and Related Work}
\label{sec:background}
The system proposed in this article combines three research areas within biometrics, \ie information fusion (subsection \ref{subsec:background_fusion}), computational workload reduction (subsection \ref{subsec:background_workload}), and template protection (subsection \ref{subsec:background_templateprotection}). This section provides a brief overview of the relevant background information and key related works in those areas. 

\subsection{Information Fusion}
\label{subsec:background_fusion}
Information fusion can be used in order to improve the discriminative power of a biometric recognition system. Referred to as ``multi-biometric systems'', they take advantage of multiple information sources which are combined (fused) in some way. Following fusion categories can be generally distinguished in the context of biometrics \cite{Ross-HandbookMultibiometrics-Springer-2006,ISO-Fusion-2015}:

\begin{LaTeXdescription}
\item[Multi-type] where multiple biometric characteristics (such as facial images and fingerprint scans) are used.
\item[Multi-sensorial] where the biometric data acquisition is conducted with diverse sensors providing complementary information (for example, near-infrared and visible-wavelength cameras).
\item[Multi-algorithm] where the biometric data is processed utilising several complementary algorithms (for instance, image descriptors based on texture and keypoint information).
\item[Multi-instance] where more than one instances of the same underlying type of biometric characteristic are used (\eg the images of right and left iris).
\item[Multi-sample] where several biometric samples stemming from one type of biometric characteristic are used (\eg multiple acquisitions of a fingerprint scan with the purpose of detecting reliable regions or assuring the quality and consistency of the acquired data).
\end{LaTeXdescription}

Information fusion can occur at different steps of the biometric processing pipeline \cite{Ross-HandbookMultibiometrics-Springer-2006,ISO-Fusion-2015}, including:

\begin{LaTeXdescription}
\item[Sensor] where raw data (\eg images acquired by different sensors or multiple samples) is combined before other processing steps \cite{Jain-FingerprintMosaicking-2002,Kusuma-PCAFusion-2011}.
\item[Feature] where the extracted feature sets (\eg from multiple samples) are consolidated \cite{Kanhangad-Fusion-2011,Yan-Fusion-2015}.
\item[Score] where the comparison scores computed through different information channels are combined (\eg averaged) \cite{Snelick-Fusion-2003,Jain-ScoreNormalization-2005}.
\item[Rank] where the orders (ranks) of potential matches between a probe and the enrolment database obtained through different information channels are consolidated \cite{Abaza-Fusion-2009,Kumar-Fusion-2011}.
\item[Decision] where the decisions (\ie acceptance or rejection) obtained through multiple information channels are combined (\eg by a majority vote) \cite{Prabhakar-Fusion-2002,Paul-Fusion-2014}.
\end{LaTeXdescription}

In the context of this work, fusion of multiple samples (from different data subjects) on feature level are of most interest, as the system proposed in section \ref{sec:proposedsystem} is designed to operate at those level of the biometric processing pipeline.

The topic of information fusion in biometrics has been addressed extensively in the scientific literature. In \cite{Ross-HandbookMultibiometrics-Springer-2006}, a general introduction to this topic is given, while \cite{ISO-Fusion-2015,Dinca-FusionSurvey-2017,Singh-Fusion-2019} provide recent and comprehensive surveys of this research area.

\subsection{Computational Workload Reduction}
\label{subsec:background_workload}
Maintaining fast biometric identification system response times often requires optimisation or additional investments as the size of the enrolment database increases. The computational costs of the typical, exhaustive search-based, retrieval method tend to grow linearly with the number of enrolled data subjects \cite{Daugman-DecisionLandscapes-UCAM-2000}. Naturally, the expansion of the underlying hardware (\eg by using many servers which facilitate distributing the computations) can be used to maintain quick system response times; however, this solution carries with it high monetary costs, such as the purchase of the equipment, its installation and maintenance, \etc While hardware investments are often inevitable, an often overlooked possibility is the optimisation of the underlying software and/or algorithms. In this context, the field of \textit{computational workload reduction} has emerged in recent years and numerous methods have been proposed which can help to mitigate some of the costs of the physical infrastructure. The goal of such methods is the reduction of the required amount of computations for some specific tasks in the biometric recognition pipeline. As the computational costs of the biometric template comparisons typically dominates the overall computational effort in biometric identification transactions, most of the approaches proposed in the literature are aimed specifically at optimising this step of the biometric identification pipeline \cite{Drozdowski-WorkloadSurvey-IET-2019}. More specifically, two broad classes of approaches can be distinguished: \textit{pre-selection}, concentrating on the reduction of the search space, \ie the number of necessary template comparisons (see \eg \cite{Kavati-SearchReductionSurvey-IGI-2018}), and \textit{feature transformation}, aimed at lowering the computational cost of the individual template comparisons (see \eg \cite{Drozdowski-DeepFaceBinarisation-ICIP-2018}). The former are of interest in the context of this article.

Numerous methods rely on the so-called pre-filtering of the enrolment database during a biometric identification transaction. Such methods depend on categorical or weakly discriminative features (\eg geographic and/or demographic metadata \cite{Gehrmann-MetadataFiltering-2019} or soft biometrics \cite{Dantcheva-SoftBiometricsSurvey-TIFS-2016}), whereby the potential search space can be narrowed down quickly prior to considering the actual highly discriminative, but more computationally expensive to compute, biometric features. Conceptually similar two or multi-stage methods operating on weakly discriminative, compact representations (\eg dimensionally-reduced or binarised) representation of biometric data have also been considered \cite{Gentile-TwoStageIris-BTAS-2009, Billeb-SpeakerTwoStage-BIOSIG-2014, Pflug-HistogramBinarisation-CYBCONF-2015}. Likewise, general concepts of coarse-to-fine search, nearest-neighbour search, and clustering based on the feature sets extracted from biometric samples have also been proposed \cite{Kavati-SearchReductionSurvey-IGI-2018,Drozdowski-WorkloadSurvey-IET-2019}.

More complex methods directly utilising the extracted biometric features and aimed at creating an intelligent search structure (\eg a search tree) have been shown to be capable of significantly reducing the computational workload. In \cite{Drozdowski-BloomFilterIndexing-IET-2018}, a tree-based indexing and retrieval system for iris data has been proposed. Many successful methods of biometric indexing integrate information fusion; for example, \cite{Iloanusi-FingerprintIndexing-PRL-2014,Drozdowski-MultiIrisIndexing-IJCB-2017} for multi-instance fingerprint and iris data, respectively. Furthermore, generic multi-biometric indexing methods have also been proposed \eg in \cite{Jayaraman-MultimodalIndexing-ICISS-2008,Gyaourova-IndexCodes-CVPRW-2009,Gyaourova-IndexCodes-TIFS-2012}. 

In \cite{Drozdowski-Kstage-2019}, a multi-biometric cascade has been proposed with the aim of successively filtering the candidate short-lists based on score-level information. Similar concepts were utilised in \cite{Drozdowski-MorphingPreselection-ICASSP-2019,Drozdowski-SignalFusionWorkload-2021}, where a signal-level fusion (\ie morphing, see \eg \cite{Scherhag-MorphingSurvey-2019}) of facial images facilitates a computationally efficient and accurate indexing and retrieval for biometric identification. Those methods are most closely related to the indexing and retrieval method presented in this article.

Generally, the methods mentioned in this subsection often require the storage of additional information (\eg metadata) and/or a kind of a ``setup'' step (\eg creation of a search structure) which requires some computational effort, but only needs to be performed infrequently. On the other hand, many of the described methods facilitate the reduction of computational workload associated with biometric identification transactions by several orders of magnitude \wrt the typical exhaustive-search based retrieval method.

\begin{table}[!ht]
\centering
\caption{Properties of template protection categories}\label{tab:overview_btp}
\resizebox{\columnwidth}{!}{
\renewcommand*{\arraystretch}{1.2}
\begin{tabular}{cccccccc}
\toprule
\begin{tabular}{@{}c@{}}\textbf{Template}\\ \textbf{protection}\\ \textbf{category}\end{tabular} & \rotatebox[origin=c]{90}{\textbf{Unlinkability}} & \rotatebox[origin=c]{90}{\textbf{\hspace{0.1 mm} Irreversibility \hspace{0.1 mm}}} & \rotatebox[origin=c]{90}{\textbf{Renewability}} & \rotatebox[origin=c]{90}{\begin{tabular}{@{}c@{}}\textbf{Performance}\\ \textbf{preservation}\end{tabular}} & \rotatebox[origin=c]{90}{\begin{tabular}{@{}c@{}}\textbf{Efficient}\\ \textbf{comparison}\end{tabular}}  &  \rotatebox[origin=c]{90}{\begin{tabular}{@{}c@{}}\textbf{Key}\\ \textbf{derivation}\end{tabular}} 
\\ \midrule
\begin{tabular}{@{}c@{}}Cancelable\\ biometrics \end{tabular} & \quad\cmark \quad \quad & \quad\cmark \quad \quad & \quad\cmark \quad \quad & (\cmark) & \cmark &  \\\midrule
\begin{tabular}{@{}c@{}}Biometric\\ cryptosystems \end{tabular}  & \cmark & \cmark & \cmark & (\cmark) & (\cmark) &  \cmark \\\midrule
\begin{tabular}{@{}c@{}}Homomorphic \\ encryption\end{tabular}  & \cmark & \cmark & \cmark & \cmark & (\cmark) & \\\bottomrule
 \end{tabular}
}
\end{table}

\begin{table*}[!ht] 
	\begin{center}
	\caption{Overview of most relevant privacy-preserving WR schemes for face-based identification systems (results reported for best configurations and scenarios; note the differences in the used evaluation datasets and performence metrics)}\label{tab:overview-secure-ide-syst}
	  \begin{adjustbox}{max width=\textwidth}
			  \begin{threeparttable}

\begin{tabular}{c c c c c } \toprule
		 \textbf{Approach} & \textbf{Workload reduction category} & \textbf{Template protection category} & \textbf{Dataset} & \textbf{Biometric performance} \\ \midrule
		
		 \multirow{2}{*}{Wang \textit{et al. }~\cite{Wang-indeference-similarity-2017}} & \multirow{2}{*}{\shortstack{Feature transformation  \\ Pre-selection }} & \multirow{2}{*}{Not traditional BTP} & \multirow{2}{*}{\shortstack{FERET \\ LFW}} & 89\% HR 	 \\

		 																				&										&		 	
																						&			
																						& 95\% HR \\ \midrule

		Murakami \textit{et al. }~\cite{Murakami-CancelableIndexing-2019} & Feature transformation & Cancelable biometrics & NIST BSSR1 SET3 & 0.1\% FRR, 0.022\% FAR \\ \midrule

		\multirow{6}{*}{Dong \textit{et al. }~\cite{Dong-face-identification-index-2020}} & \multirow{6}{*}{Feature transformation}	&
		\multirow{6}{*}{Cancelable biometrics} & LFW (closed-set)  &  99.75\% RR-1 \\
																				&
																				&
																				&
																	LFW (open-set) 			& 97.99\% DIR, 1\% FAR 	

																							\\
																				&					&   &
																			 VGG2 (closed-set) 	&    99.03\% RR-1							
																				\\
																				&  &
																				&
																				 VGG2 (open-set) & 96.03\% DIR, 1\% FAR      										\\
															   
															   & 																																&					&	  																
															  IJB-C (closed-set)   &	80.57\% RR-1												\\
															   
															   &																																&					&																		
															 IJB-C (open-set)  &	56.80\% DIR, 1\% FAR  \\	\midrule

	\multirow{4}{*}{Sardar \textit{et al. }~\cite{Sardar-novel-cancelable-face-hashing-2020}} & \multirow{4}{*}{Feature transformation}	& \multirow{4}{*}{Cancelable biometrics} & \multirow{4}{*}{\shortstack{CASIA-V5\\IITK\\CVL\\FERET }} & 99.85\% CRR-1 \\

																   &																		&															&					&		100\% CRR-1			\\				
															   
															   &								&				&								& 100\% CRR-1		\\	
															   
															   &								&								&								&	 100\% CRR-1							\\ \midrule
Drozdowski \etal~\cite{Drozdowski-HomomorphicIdentificationFace-BIOSIG-2019} 	& Feature transformation & Homomorphic encryption	& FERET & $\sim$5\% FNIR, 1\% FPIR \\					   \midrule

	Engelsma \textit{et al. }~\cite{Engelsma-HomomorphicIdentification-2020} & Feature transformation & Homomorphic encryption	& MegaFace & 81.4\% RR-1 \\    \midrule
	
	\multirow{3}{*}{Osorio-Roig \textit{et al. }~\cite{OsorioRoig-StableHashFaceIdentification-TBIOM-2021}} & \multirow{3}{*}{\shortstack{Pre-selection }}	& \multirow{3}{*}{\shortstack{Homomorphic encryption}} & \multirow{3}{*}{\shortstack{FEI\\FERET\\LFW}} & \multirow{3}{*}{\shortstack{0.0\% FPIR, 0.0\% FNIR\\0.0\% FPIR, 0.2\% FNIR\\1.0\% FPIR, 2.5\% FNIR}} \\
	& & & & \\
	& & & & \\
	\bottomrule																																														   																												  
																  
\end{tabular} 	

\begin{tablenotes}
  \item HR: Hit Rate
  \item FRR: False Rejection Rate
  \item FAR: False Acceptance Rate
  \item RR-1: Rank-1 Identification Rate
  \item DIR: Detection and Identification Rate
  \item CRR: Correct Recognition Rate at Rank-1 
  \item FPIR: False Positive Identification Rate
  \item FNIR: False Negative Identification Rate
\end{tablenotes}
\end{threeparttable} 
\end{adjustbox}
\end{center}
\end{table*}

\subsection{Biometric Template Protection}
\label{subsec:background_templateprotection}
Biometric template protection represents an active field of research since more than two decades. Comprehensive surveys on this topic can be found in \cite{5396650,Rathgeb-TemplateProtection-EURASIP-2011,ISO11-TemplateProtection,Nandakumar-TemplateProtection-IEEE-2015}. Biometric template protection methods are usually categorised as \emph{cancelable biometrics} and \emph{biometric cryptosystems}. Cancelable biometrics employ transforms in signal or feature domain which enable a biometric comparison in the transformed (encrypted) domain \cite{Patel-CancelableBiometrics-2015}. Biometric cryptosystems commonly bind a key to a biometric feature vector resulting in a protected template. Biometric comparison is then performed indirectly by verifying the correctness of a retrieved key \cite{BUludag04a}. Further, homomorphic encryption can be employed for biometric template protection \cite{Aguilar-Homomorphic-2013}. Homomorphic encryption makes it possible to compute operations in the encrypted domain which are functionally equivalent to those in the plaintext domain and thus enables the estimation of certain distances between homomorphically encrypted biometric templates. As defined in ISO/IEC IS 24745 \cite{ISO11-TemplateProtection}, biometric template protection schemes shall fulfil the following requirements:

\begin{LaTeXdescription}
\item[Unlinkability] the infeasibility of determining if two or more protected templates were derived from the same biometric instance, \eg face. By fulfilling this property, cross-matching across different databases is prevented. 
\item[Irreversibility] the infeasibility of reconstructing the original biometric data given a protected template and its corresponding auxiliary data. With this property fulfilled, the privacy of the users' data is increased, and additionally the security of the system is increased against presentation and replay attacks. Depending on the used template protection method, guaranteeing this property may rely on sufficiently protecting a certain secret (\eg private encryption key(s)) from being compromised by an attacker.
\item[Renewability] the possibility of revoking old protected templates and creating new ones from the same biometric instance and/or sample, \eg face image. With this property fulfilled, it is possible to revoke and reissue the templates in case the database is compromised, thereby preventing misuse.
\item[Performance preservation] the requirement of the biometric performance not being significantly impaired by the protection scheme.
\end{LaTeXdescription}

Table \ref{tab:overview_btp} lists the mentioned types of biometric template protection and their properties \wrt the above criteria as well as key derivation and efficient biometric comparison. The majority of approaches on cancelable biometrics and biometric cryptosystems report a performance gap between protected and original (unprotected) systems \cite{Nandakumar-TemplateProtection-IEEE-2015}, as opposed to approaches employing homomorphic encryption. Cancelable biometrics usually employ a biometric comparator similar or equal to that of unprotected biometric systems. Therefore, cancelable biometrics are expected to maintain the comparison speed of the unprotected system which makes them also suitable for biometric identification \cite{Drozdowski-WorkloadSurvey-IET-2019}. In contrast, biometric cryptosystems may need more complex comparators. Similarly, homomorphic encryption usually requires higher computational effort. Practical applications of certain template protection methods, \eg homomorphic encryption, rely on maintaining the secrecy of the private key(s) used to protect the data (see also subsection \ref{subsec:results_security}).

Some research efforts and standardisation activities have been devoted to establishing metrics for evaluating the aforementioned properties of biometric template protection schemes, \eg in \cite{Nagar10a,Wang12a,Gomez-Barrero-FrameworkForUnlinkability-TemplateProtection-2018,ISO-IEC-30136}. Nonetheless, additional specific cryptanalytic methods may be necessary to precisely estimate the security/privacy protection achieved by a particular template protection scheme. Moreover, the result of such an evaluation also depends on the biometric data to which the template protection system is applied. This makes a comparison of published results difficult and sometimes misleading.

In 2001, Ratha \etal \cite{BRatha01a} proposed the first cancelable face recognition system using image warping to transform biometric data in the image domain. Another popular cancelable transformation of face images based on random convolution kernels was presented in \cite{Savvides04a}. In contrast to \cite{BRatha01a}, this approach employs a fundamentally reversible distortion of the biometric signal based on some random seed which later coined the term ``biometric salting''. The majority of published cancelable face recognition schemes applies transformations in the feature domain \cite{Patel-CancelableBiometrics-2015}. Over the past years, numerous feature transformations have been proposed in order to construct face-based cancelable biometrics, \eg BioHashing \cite{Teoh06}, BioTokens \cite{Boult06}, and Bloom filters \cite{GomezBarrero2016}. Recently, feature transformations have been specifically designed for deep convolutional neural networks, \eg random subnetwork selection \cite{Mai21}. Analyses of some popular cancelable face recognition systems have uncovered security gaps, \eg in  \cite{KONG20061359,BRINGER2017239,Ghammam20,Kirchgasser20b}, and already led or are expected to lead to (continuous) improvements of such schemes. 
\begin{figure*}[!ht]
\centering
\includegraphics[width=0.9\textwidth]{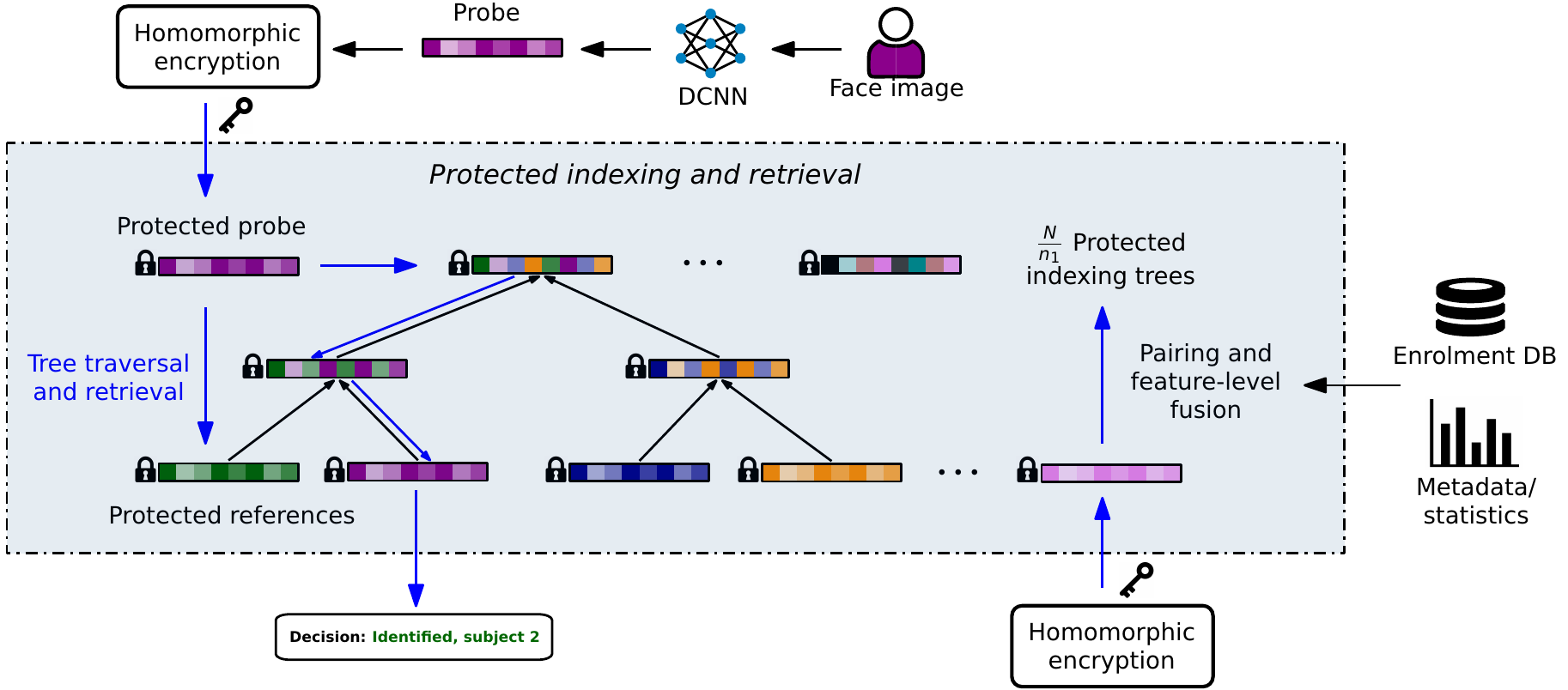}
\caption{Conceptual overview of database indexing and retrieval in the proposed system}
\label{fig:overview}
\end{figure*} 
Regarding biometric cryptosystems, the fuzzy commitment scheme \cite{BJuels99a} and the fuzzy vault scheme \cite{BJuels02a} represent widely used cryptographic primitives. Both schemes enable an error-tolerant protection of (biometric) data by binding them with a secret, \ie key. Binarised face feature vectors have been protected through the fuzzy commitment scheme in various scientific publications, \eg in \cite{vanderVeen06,Ao09}. Also some works have employed the fuzzy vault scheme for face template protection, \eg in \cite{Frassen08,Wang07,rathgeb2021deep}. It is worth mentioning that some template protection approaches combine concepts of cancelable biometrics with those of biometric cryptosystems resulting in hybrid schemes \cite{Rathgeb-TemplateProtection-EURASIP-2011}.

For a long time, homomorphic encryption has been considered as impractical for biometric template protection due to its computational workload. However, in the last years, homomorphic encryption has been applied effectively to face-based verification where practical processing times could be achieved on commodity hardware \cite{Boddeti-HE-2019}. Depending on the used homomorphic cryptosystem, different feature type transformations might be required \cite{Kolberg-FaceHE-BTP-BIOSIG-2020}.

Relevant works on biometric template protection for face-based identification systems, \ie one-to-many comparisons, are shown in table \ref{tab:overview-secure-ide-syst}. Some of listed approaches are cancelable biometrics which usually retain the biometric comparator of the corresponding unprotected system. As mentioned earlier, this property makes these approaches well suited to be applied in identification mode. In addition, approaches for face identification with homomorphic encryption have been proposed, \eg in \cite{Drozdowski-HomomorphicIdentificationFace-BIOSIG-2019,Engelsma-HomomorphicIdentification-2020}. These works, use different concepts to maximize the efficiency of homomorphic encryption, including optimisation strategies, \eg batching or dimensionality reduction. In summary, it is important to note that all published works on biometric template protection for face identification employ an exhaustive search, \ie these scheme scale linearly \wrt to the number of protected reference face templates in the database.

As mentioned in subsection \ref{subsec:background_templateprotection}, a large concern in biometric system deployments is the risk of data exposure\footnote{This risk is not merely hypothetical -- consider real hacks such as those described in \cite{OPM-DataHackNews-2015}.}. Simultaneous efficient indexing and protection of biometric data has been proposed \eg in \cite{Li-ProtectedIndexing-IEEE-2016,Drozdowski-ProtectedIrisIndexing-EUSIPCO-2018,Wang-indeference-similarity-2017,Murakami-ProtectedIndexing-2019,Murakami-CancelableIndexing-2019}. \cite{Drozdowski-HomomorphicIdentificationFace-BIOSIG-2019,Kolberg-FaceHE-BTP-BIOSIG-2020,Engelsma-HomomorphicIdentification-2020} explore the use of homomorphic encryption in conjunction with biometric identification and attempt to reduce computational workload by applying a packing strategy to decrease the computation between the ciphertexts or by applying different (more computationally efficient) HE schemes. In summary, coupling biometric template protection with computational workload reduction (\ie ensuring privacy and computational efficiency in addition to high biometric performance) is an insufficiently addressed topic in biometric research.

\section{Proposed System}
\label{sec:proposedsystem}
The high-level, conceptual overview of the proposed system is demonstrated in figure \ref{fig:overview} (indexing) and algorithm \ref{alg:retrieval} (retrieval). The proposed system relies on creation of an efficient tree-like search structure by fusing the reference templates stored in the enrolment database. Let $N$ be the number of subjects in the enrolment database and $n_{i}$ (selected from the set $\set{2^{x} \mid x \in \mathbb{N^{+}}}$) be the number of subjects contributing to the fused templates at the $i$'th level of the tree-like search structure. For instance, in figure \ref{fig:overview}, the roots of the indexing trees consist of four subjects, \ie $n_{1} = 4$. On the following levels, this number decreases ($n_{2} = 2$) until non-fused reference templates are considered at the final level ($n_{3} = 1$). Subsections \ref{subsec:proposedsystem_pairselection} and \ref{subsec:proposedsystem_fusion} provide details on how the templates to be fused are paired and what information fusion methods are used. During a biometric identification transaction, the created search structure is traversed whereby the biometric probe is compared against the fused templates in order to successively narrow down the list of potential candidate identities at each level of the search structure. The search structure has $\log_{2} n_{1} + 1$ levels; let $k_{i}$ represent the fraction of nodes and their corresponding identities selected at the $i$'th level of the search structure. The key idea here is for $k$ to be relatively small and decreasing at each level of the search structure. Subsection \ref{subsec:proposedsystem_retrieval} provides more details on this retrieval algorithm, as well as a theoretical analysis of the possible gains in computational efficiency \wrt a na{\"i}ve exhaustive search-based retrieval algorithm. The proposed system also allows for a seamless integration of template protection as described in subsection \ref{subsec:proposedsystem_templateprotection}.

\begin{algorithm}[!ht]
\caption{Retrieval in the proposed system}
\label{alg:retrieval}
\renewcommand{\algorithmicrequire}{\textbf{Input:}}
\renewcommand{\algorithmicensure}{\textbf{Output:}}
\begin{algorithmic}[1]
 \REQUIRE $probe, indexing\_trees$
 \ENSURE $candidates$
 \STATE $candidates$ $\gets$ roots of $indexing\_trees$
 \FOR{$i = 1$ to $\log_{2} n_{1} + 1$}
 	\STATE $scores$ $\gets$ compare $probe$ with all $candidates$
 	\STATE $best\_scores$ $\gets$ find $\left\Vert scores \right\Vert \cdot k_{i}$ highest $scores$
 	\STATE $candidates$ $\gets$ select $candidates$ with $best\_scores$
 \ENDFOR
 \RETURN $candidates$
\end{algorithmic}
\end{algorithm}

\subsection{Retrieval}
\label{subsec:proposedsystem_retrieval}
Since the fused templates retain sufficient discriminative power, the probes exhibit better comparison scores against their respective correct (mated) fused templates than against the other (non-mated) fused templates. Consequently, it is possible to make a robust pre-selection of a candidate short-list to be passed onto the next level of the cascade. In a successive manner, which is conceptually similar to the previous works on multi-modal and signal-level fusion-based cascades of Drozdowski \etal \cite{Drozdowski-Kstage-2019,Drozdowski-SignalFusionWorkload-2021}, the candidate short-list shrinks at each level, thus resulting in fewer template comparisons being made and hence in computational workload reduction. The computational workload ($W$) \cite{ISO-PerformanceReporting-2021} of the proposed retrieval scenario can be obtained using the following formula: 

\begin{equation}
\label{eq:workload}
W = \frac{\frac{N}{n_{1}} + \sum_{l=2}^{\log_{2} n_{1}} 2k_{l}}{N} \cdot 100 \%
\end{equation}

This equation expresses the computational workload of the proposed indexing and retrieval method as a percentage of the workload required in the typical baseline scenario where an exhaustive ($1$:$N$) search is carried out. 

\begin{figure}[!ht]
\centering
\includegraphics[width=\columnwidth]{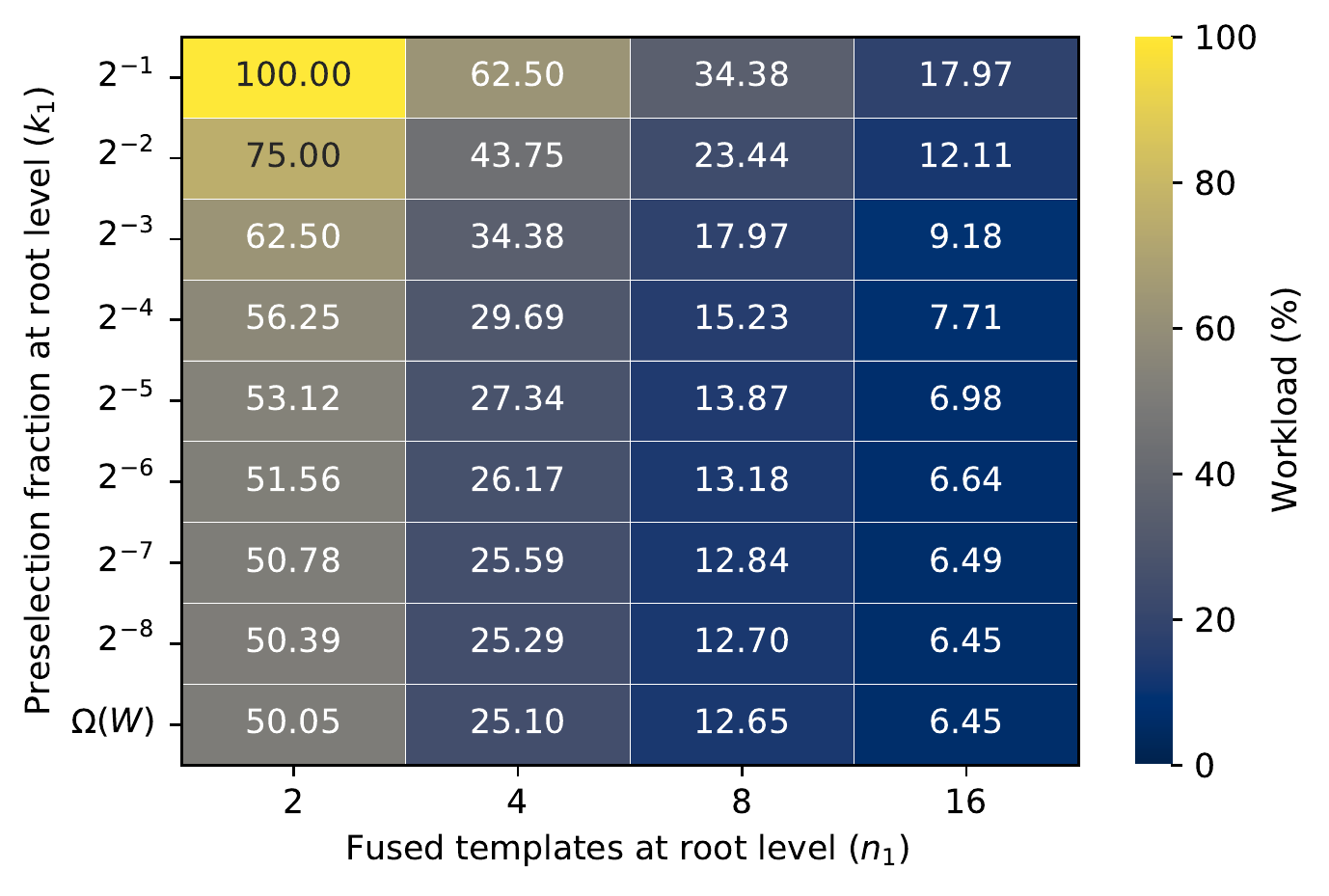}
\caption{Theoretical overview of computational workload depending on the parameters of the proposed system}
\label{fig:workload_theoretical}
\end{figure}

Figure \ref{fig:workload_theoretical} illustrates the impact of the parameters of the proposed system on its computational workload. The x-axis shows the number of fused templates at the root level of the search tree, \ie how many templates are fused with each other (the $n_{1}$ parameter). The y-axis denotes the fraction of templates pre-selected at the root level ($k_{1}$) followed by a cascade with logarithmically decreasing pre-selection sizes. $\Omega(\text{W})$ denotes the theoretical lower bound, \ie $1$ fused template being pre-selected at each level of the cascade. The values in the figure are given for an $N$ value which was used in the empirical experiments reported later on in the paper. With an increasing $N$ value (\ie growth of enrolment database size), the lower bound of computational workload for the proposed system can be expressed as follows:

\begin{equation}
\label{eq:workload_lower_bound}
\lim\limits_{N \to \infty} \Omega(\text{W}) = \frac{1}{n_{1}} \cdot 100 \% 
\end{equation}

Following three observations can be made:
\begin{enumerate}
\item There do exist configurations (based on $k_{1}$ and $n_{1}$ parameters) which require significantly less computational workload than the baseline. In other words, provided sufficient discriminative power, the proposed system is capable of reducing the computational workload in biometric identification.
\item The $k_{1}$ parameter has a moderate impact on the possible computational workload reduction within each of the four columns. Indeed, diminishing returns are quickly approached as $k_{1}$ at root level decreases -- \cf the workload at lower bound with values at \eg $k_{1} = 2^{-4}$ or $k_{1} = 2^{-5}$.
\item The $n_{1}$ parameter has a large impact on the possible computational workload reduction. Each time $n_{1}$ is doubled (\ie the height of the cascade increases), the workload is approximately halved for all $k_{1}$ values. 
\end{enumerate}

From the above observations, it follows that for a workload-centric perspective, one would prefer as high $n_{1}$ value as possible in order to achieve highest possible workload reduction; the $k_{1}$ values usually being a secondary concern. Consider, for instance, that for $n_{1} = 8$, the lower bound for achievable workload is $12.65\%$. If $n_{1}$ can be increased to $16$, the aforementioned workload is achieved already at a relatively high fraction ($k_{1} = 2^{-2}$) of templates being pre-selected at that level. It is, however, important to remember that the indexing and pre-selection may increase the false-negative errors, \ie the parameters $n_{1}$ and $k_{1}$ likely cannot be set to achieve the lower bound for computational workload without simultaneously causing a significant impairment of the biometric performance. In other words, the desired reduction in computational workload needs to be feasible \wrt the discriminative power of the utilised recognition system. This trade-off between computational workload and biometric performance is evaluated empirically later on in this article.

\subsection{Selection of Feature Vector Pairs}
\label{subsec:proposedsystem_pairselection}
Deciding which parent samples to fuse with each other is expected to have a non-trivial impact on the efficacy of the proposed system. With an intelligent matching of the fused subject pairs, an increase in the discriminative power of the pre-selection procedure is expected, thereby improving the overall results of the proposed system in terms of biometric performance and computational workload.

Ideally, similar data subjects/samples would be fused with each other. Conceptually, matching such pairs belongs to an old and well-known class of combinatorial optimisation problems. One could formulate it in terms of a stable roommates or stable marriage problem. In practical experiments, however, such formulation has been plagued by issues related to ``odd pairs'' and solvability on a large set of data (see \cite{Pittel-StableRoommates-1994,Chung-StableRoommates-2000,Rottcher-MorphingDoppelganger-2020}). In this work, those issues are circumvented by optimising the matching algorithm with a global cost function instead of seeking a stable matching. The benefit of this approach is that some poorly matched (\ie with a high cost) pairs are allowed, while the overall matchings are well-optimised for a given enrolment database. In practical experiments, this formulation (corresponding to the assignment problem) has been applied successfully.

More formally, let $S$ represent the set of data subjects present in the enrolment database. A bijective mapping of this set to itself is sought, \ie $f: S \to S$, with an additional constraint that the subjects may not be mapped to themselves, \ie $\forall s \in S, f(s) \neq s$. Given a weight function $C: S \times S \to \mathbb{R}^{+}$, the aim of a successful mapping is to minimise $\displaystyle\sum_{s \in S} C_{s, f(s)}$. This work considered three methods for mapping selection:

\begin{LaTeXdescription}
\item[Random] samples are paired purely by chance, \ie no special algorithm is used for the pair selection.
\item[Soft-biometric] similarity based on soft-biometric attributes (sex, race, age) is computed between the enrolled samples as a basis for the assignment.
\item[Similarity-score] similarity based on non-mated comparison scores between the enrolled samples computed with a facial recognition system serves as a basis for the assignment.
\end{LaTeXdescription}

In practice, given an $N$-subject large enrolment database, a square matrix with the aforementioned similarity scores (soft-biometric or recognition based) can be created as illustrated in equation \ref{eq:scores_matrix}. There, $S_{x}$ denotes the $x$'th data subject, while $c_{x,y}$ denotes the cost of pairing the $x$'th and $y$'th data subject with each other. To represent the constraint of data subjects not being allowed to be paired with themselves, the diagonal is set to $\infty$. In the concrete software implementation, the largest possible value of a floating-point datatype is used instead.

\begin{equation}
\label{eq:scores_matrix}
C = \bordermatrix{~ & S_{1} & S_{2} & S_{3} & \cdots & S_{N} \cr
                  S_{1} & \infty & c_{1,2} & c_{1,3} & \cdots & c_{1,N} \cr
                  S_{2} & c_{2,1} & \infty & c_{2,3} & \cdots & c_{2,N} \cr
                  S_{3} & c_{3,1} & c_{3,2} & \infty & \cdots & c_{3,N} \cr
                  \vdots & \vdots & \vdots & \vdots & \ddots & \vdots \cr
                  S_{N} & c_{N,1} & c_{N,2} & c_{N,3} & \cdots & \infty \cr
                  }
\end{equation}

As formulated above, a polynomial time solution for the problem exists using the so-called Hungarian algorithm \cite{Kuhn-HungarianAlgorithm-1955}. An iterative procedure is used to produce pairs for subsequent steps in the cascade, \ie for $n > 2$. While computationally intensive, this step is only required once (offline) during indexing of the enrolment database and not during every retrieval.

\begin{figure}[!ht]
\def\currentsize{0.225}
\setlength{\tabcolsep}{1pt}
\subfloat[Subject]{
\begin{tabular}{c}
\footnotesize \\
\includegraphics[width=\currentsize\columnwidth]{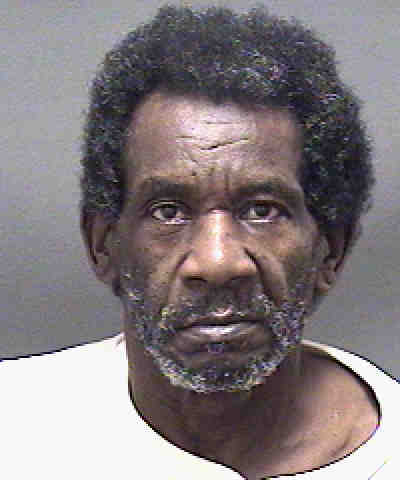} \\
\includegraphics[width=\currentsize\columnwidth]{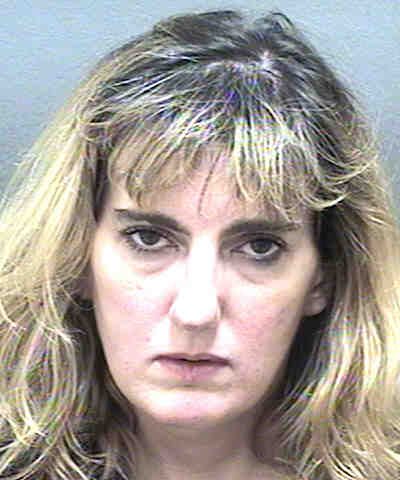}
\end{tabular}
}
\subfloat[Selected paired subject]{
\begin{tabular}{ccc}
\footnotesize Soft-biometric & \footnotesize Similarity-score 1 & \footnotesize Similarity-score 2 \\
\includegraphics[width=\currentsize\columnwidth]{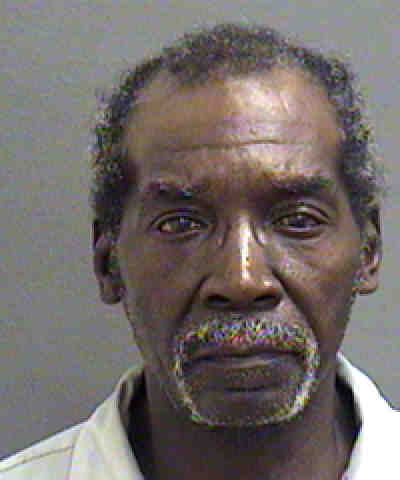} &
\includegraphics[width=\currentsize\columnwidth]{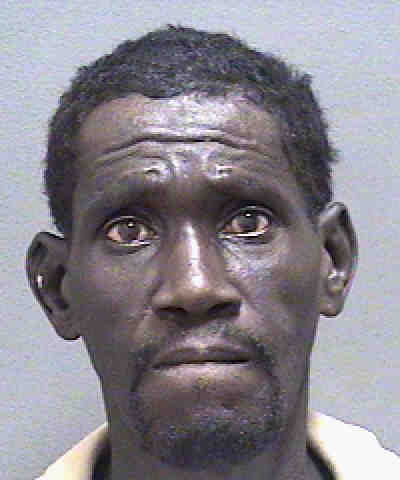} &
\includegraphics[width=\currentsize\columnwidth]{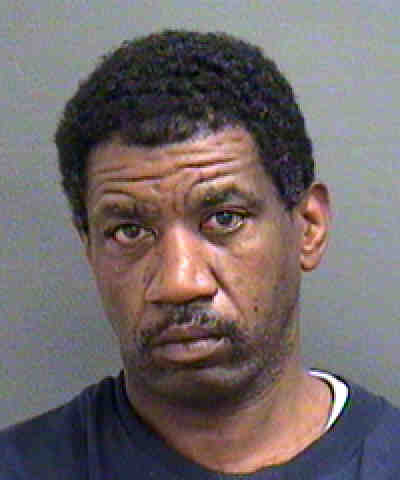} \\
\includegraphics[width=\currentsize\columnwidth]{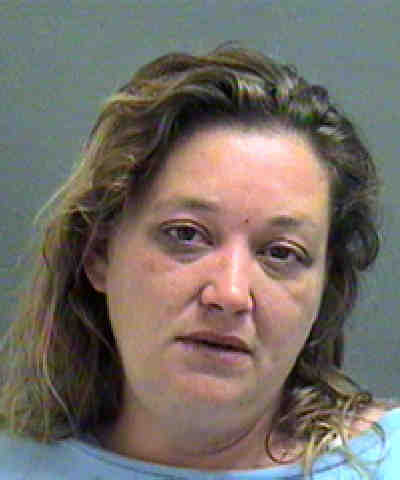} &
\includegraphics[width=\currentsize\columnwidth]{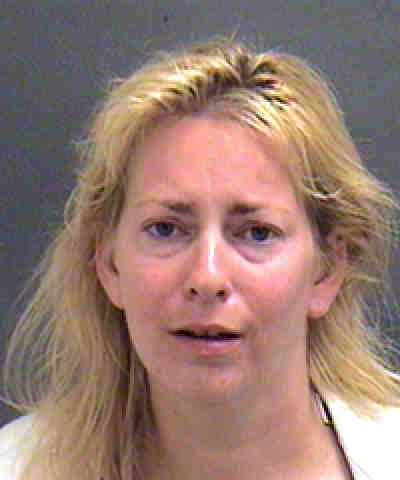} &
\includegraphics[width=\currentsize\columnwidth]{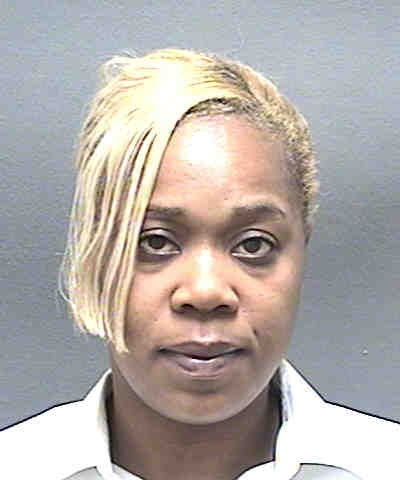}
\end{tabular}
}
\caption{Example images of pairings found using the proposed method}
\label{fig:example_pairs}
\end{figure}

Figure \ref{fig:example_pairs} shows examples of subjects paired using the soft-biometric and similarity-score based methods described above. Details on the dataset and face recognition systems used in the experiments are provided in section \ref{sec:experimentalsetup}.

\subsection{Feature Fusion Methods}
\label{subsec:proposedsystem_fusion}
The choice of information fusion method has a non-trivial impact on the discriminative power and hence the biometric performance of a recognition system \cite{Dinca-FusionSurvey-2017,Singh-Fusion-2019}. The overarching goal of feature-level fusion is to create a fused feature vector $\overline{\mathbf{v}}=(\overline{v}_i)^n_{i=1}$, $\overline{v}_i \in \mathbb{R}$ from a pair of feature vectors $\mathbf{v}$ and $\mathbf{v}'$ of same size. For simplicity, this definition and the formulas of the used fusion methods are provided in a notation for two feature vectors being fused (\ie $n = 2$). However, for the specific application scenario considered in this article, they can be (and are) trivially extended to an arbitrary number ($n$) of feature vectors from the set $\{2^{i} \mid i \in \mathbb{N}^{+}\}$. Three fundamentally different types of feature fusion methods are considered and described below. In the provided formulas, $\mu$ represents an overall average value of elements at a given position and is computed on a disjoint training set of feature vectors.

\subsubsection{Average-based}
\label{subsubsec:proposedsystem_fusion_average}
An intuitive method to fuse feature vectors is averaging. Following variants of this method are considered:

\begin{LaTeXdescription}
\item[Simple average] The arithmetic mean of the elements at each feature position is taken: $\overline{v_i}=(v_i+ v'_i)/2$.
\item[Weighted average] The arithmetic mean of the elements at each feature position is taken and additionally weighted by the distance of this element from an overall average at the given position computed on a training set:
$\overline{v_i} = (v_i  |v_i-\mu_i| + v'_i  |v'_i-\mu_i|) / 2$. In other words, elements which strongly deviate from the average are assigned more weight.
\end{LaTeXdescription}

The above two methods are henceforth referred to as ``Average-1'' and ``Average-2''.

\subsubsection{Distance-based}
\label{subsubsec:proposedsystem_fusion_distance}
The following methods rely on putting the values of the individual elements in relation to some overall properties of the feature vectors.

\begin{LaTeXdescription}
\item[Distance from mean] For each element position, the element furthest from an overall average at the given position is chosen: $\overline{v_i} = \begin{cases}
v_i & |v_i-\mu_i| \geq |v'_i-\mu_i| \\
v'_i & \, \text{otherwise.}
\end{cases}$. In other words, the element which exhibits the strongest deviation from the average at a given position is used directly.
\item[Distance from mean rank-based] We define $\#(v_i)\in \mathbb{N}$ to be the rank of $v_i$ in the sequence of elements of $\mathbf{v}$ sorted in ascending order according to their distance to the mean $\mu$, preserving duplicate elements. Following this operation, the element with the highest rank at a given position is chosen: $\overline{v_i} = \begin{cases}
v_i & \#(v_i) \geq \#(v'_i) \\
v'_i & \, \text{otherwise.}
\end{cases}$
\end{LaTeXdescription}

The above two methods are henceforth referred to as ``Distance-1'' and ``Distance-2''.

\subsubsection{Index-based}
\label{subsubsec:proposedsystem_fusion_index}
The following methods depend on the position of the elements in the feature vectors.

\begin{LaTeXdescription}
\item[Section segregation] A portion (\eg half) of each of the contributing vectors is taken directly:
$\overline{v_i} = \begin{cases}
v_i & i \leq n/2 \\
v'_i & \, \text{otherwise.}
\end{cases}$
\item[Alternating index] The feature elements are directly taken from each of the contributing vectors in an alternating manner: $\overline{v_i} = \begin{cases}
v_i & i \equiv 1 \mod 2 \\
v'_i & \, \text{otherwise.}
\end{cases}$
\end{LaTeXdescription}

The above two methods are henceforth referred to as ``Index-1'' and ``Index-2''.

\subsection{Template Protection}
\label{subsec:proposedsystem_templateprotection}
In the proposed scheme, template protection is facilitated through integration of homomorphic encryption. In general, an encryption algorithm $E$ has the homomorphic property for an operation $\odot$ if it holds $E(m_1) \odot E(m_2) = E(m_1 \odot m_2), \forall m_1,m_2 \in M$, where $M$ is the set of all possible messages. For more details on this topic, see \eg a detailed survey in \cite{Acar-HESurvey-2018}. As shown in \cite{Drozdowski-HomomorphicIdentificationFace-BIOSIG-2019,Kolberg-FaceHE-BTP-BIOSIG-2020}, the template comparator for biometric templates extracted from facial images can be feasibly implemented in the homomorphically encrypted domain. In other words, during template comparison in the protected domain, operations which are mathematically identical to those in the unprotected domain are conducted. Thus, the protection of the templates in the proposed scheme results in no loss of biometric performance (in contrast to typical biometric cryptosystems and cancelable biometrics). Using homomorphic encryption libraries described in subsection \ref{subsec:experimentalsetup_homomorphicencryption}, the biometric probes, as well as the stored biometric reference templates and the index constructed from the fused templates can all be encrypted and compared in the protected domain, thereby fulfilling the biometric template protection objectives (unlinkability, irreversibility, renewability, and performance preservation) of ISO/IEC IS 24745 \cite{ISO11-TemplateProtection} (see subsection \ref{subsec:results_security} for more details).

\section{Experimental Setup}
\label{sec:experimentalsetup}
This section provides a detailed description of the setup for the conducted experiments. The used dataset and face recognition systems are described in subsections \ref{subsec:experimentalsetup_dataset} and \ref{subsec:experimentalsetup_facerecognitionsystems}, respectively; subsection \ref{subsec:experimentalsetup_homomorphicencryption} describes the used homomorphic encryption software, while subsection \ref{subsec:experimentalsetup_evaluationmetrics} gives an overview of the evaluation methodology and metrics.

\subsection{Dataset}
\label{subsec:experimentalsetup_dataset}
\begin{figure}[!ht]
\centering
\includegraphics[width=0.24\columnwidth]{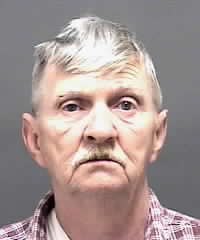} \hfill
\includegraphics[width=0.24\columnwidth]{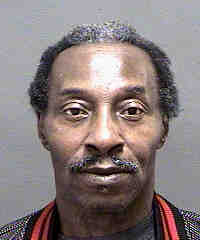} \hfill
\includegraphics[width=0.24\columnwidth]{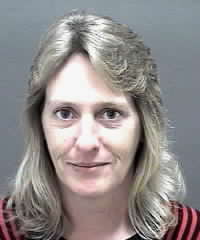} \hfill
\includegraphics[width=0.24\columnwidth]{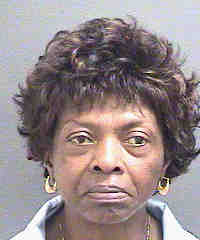}
\caption{Example images from the used dataset}
\label{fig:example_images_dataset}
\end{figure}

The academic MORPH dataset by Ricanek \etal \cite{Ricanek-MORPH-2006} has been used in the experiments. A subset of images was selected based on approximate conformance with ICAO requirements for passport images \cite{ICAO-9303-p9-2015}. As the proposed system is aimed to function with such semi-constrained images, the so-called ``in-the-wild'' datasets were not considered. Furthermore, the chosen dataset facilitates the soft-biometric pairing method described in subsection \ref{subsec:proposedsystem_pairselection}, as groundtruth metadata is available for three demographic attributes -- sex, race, and age. Figure \ref{fig:example_images_dataset} shows example images from the used dataset, while table \ref{table:dataset} provides a numerical summary of its partitioning for the experiments. 

\begin{table}[!ht]
\centering
\caption{Overview of the used dataset}
\label{table:dataset}
\resizebox{0.75\columnwidth}{!}{
\begin{tabular}{lrr}
\toprule
\textbf{Partition} & \textbf{Subjects} & \textbf{Samples} \\
\midrule
Reference & 4,096 & 4,096 \\
Probe (enrolled) & 4,096 & 12,939 \\
Probe (non-enrolled) & 1,935 & 7,123 \\
\bottomrule
\end{tabular}
}
\end{table}

\subsection{Face Recognition Systems}
\label{subsec:experimentalsetup_facerecognitionsystems}
Two well-known open-source face recognition systems are used in the experiments:

\begin{LaTeXdescription}
\item[ArcFace] A somewhat recent (initial publication in 2018), but continually improved and refined system published by Deng \etal \cite{Deng-ArcFace-2019}. The code and pre-trained model ``LResNet100E-IR,ArcFace@ms1m-refine-v2'' provided by the authors are used\footnote{\url{https://github.com/deepinsight/insightface}}.
\item[CurricularFace] A very recent system (2020) published by Huang \etal \cite{Huang-CurricularFace-2020}. The code and pre-trained model ``IR101'' provided by the authors are used\footnote{\url{https://github.com/HuangYG123/CurricularFace}}.
\end{LaTeXdescription}

Both systems achieve excellent biometric performance in popular large-scale face recognition benchmarks. The systems extract compact feature vectors with 512 floating-point elements. Those vectors can be seamlessly fused using the methods described in subsection \ref{subsec:proposedsystem_fusion}. Euclidean distance is used to compute the dissimilarity between two feature vectors.

\begin{table*}[!ht]
\centering
\caption{Summary of the used homomorphic encryption schemes}
\label{table:he_schemes}
\resizebox{0.75\textwidth}{!}{
\begin{tabular}{llrrrrr}
\toprule
\multirow{2}{*}{\textbf{Scheme}} & \multirow{2}{*}{\textbf{Data type}} & \multicolumn{3}{c}{\textbf{Execution time}} & \multicolumn{2}{c}{\textbf{Storage}} \\ \cmidrule(r){3-5} \cmidrule(l){6-7}
& & \textbf{Key generation} & \textbf{Encryption/Decryption} & \textbf{Comparison} & \textbf{Keys} & \textbf{Template} \\
\midrule
 CKKS \cite{Cheon-CKKS-2017} & float & {$\sim$}779 ms & {$\sim$}6 ms & {$\sim$}3391 ms & {$\sim$}99 MB & {$\sim$}516 KB \\
BFV \cite{DBLP:journals/iacr/FanV12} & integer & {$\sim$}255 ms & {$\sim$}76 ms & {$\sim$}618 ms & {$\sim$}12 MB & {$\sim$}132 KB \\
NTRU \cite{hoffstein1998ntru} & binary & {$\sim$}362 ms & {$\sim$}27 ms & {$\sim$}23 ms & {$\sim$}6 KB & {$\sim$}5.5 KB \\
\bottomrule
\end{tabular}
}
\end{table*}

\subsection{Homomorphic Encryption}
\label{subsec:experimentalsetup_homomorphicencryption}
Table \ref{table:he_schemes} summarises the HE schemes used to encrypt the biometric templates. Open-source implementations were used -- \cite{sealcrypto} for CKKS and BFV, and \cite{Kolberg-NTRU-2019} for NTRU. To facilitate the encryption schemes which operate using integer or binary input data, template quantisation and binarisation methods of \cite{Drozdowski-DeepFaceBinarisation-ICIP-2018} were used. The approximate execution times given in table \ref{table:he_schemes} (medians over multiple runs) refer to single operations, \eg a template comparison. Additionally, in section \ref{sec:results}, this benchmark is reported for an entire identification transaction in the baseline and proposed system. The timing benchmark was conducted on a freshly installed Linux Debian 10 on a commodity notebook running an Intel Core i7 2.7 GHz CPU and 16 GB DDR4 RAM. It should be noted that given stronger hardware (\cf \cite{Engelsma-HomomorphicIdentification-2020}), significantly faster execution times for the basic operations in the homomorphically encrypted domain can be achieved. Additionally, depending on the used HE scheme, it might be theoretically possible to incorporate acceleration of linear algebra operations in the encrypted domain \cite{cryptoeprint:2020:1483,Zong-HE-2021}. 

\subsection{Evaluation Metrics}
\label{subsec:experimentalsetup_evaluationmetrics}
In the experiments, the proposed method is evaluated against an exhaustive-search based baseline. Two key aspects are considered using standardised methods and metrics \cite{ISO-PerformanceReporting-2021} supported by additional ones which are commonly reported in the scientific literature:

\begin{LaTeXdescription}
\item[Biometric performance] In closed-set identification experiments, the CMC curves, (true-positive) identification rate (IR), and rank-1 recognition rate (RR-1) are reported. In open-set identification experiments, the DET curves and false negative identification rate at a decision threshold corresponding to a fixed false positive identification rate of 0.1\% (denoted $\mathbf{FNIR_{1000}}$).
\item[Computational workload] the overall computational workload (denoted $W$) of a single biometric identification transaction is calculated the workload reduction by the proposed scheme \wrt baseline is computed. This is done based on the necessary number of template comparisons and reported for the proposed system in percentage terms in relation to the exhaustive search-based baseline (\ie with $W = 100\%$).
\end{LaTeXdescription}

\section{Results}
\label{sec:results}
The proposed system is evaluated experimentally as follows: in subsections \ref{subsec:results_workload} and \ref{subsec:results_configs}, an analysis is conducted to establish suitable configurations which simultaneously minimise the computational workload and maximise the biometric performance. In subsection \ref{subsec:results_overall}, the overall results for the selected optimal configurations are reported. The scalability of the proposed method is briefly discussed in subsection \ref{subsec:results_scalability}.

\subsection{Analysis of Computational Workload}
\label{subsec:results_workload}
\begin{figure}[!ht]
\centering
\includegraphics[width=\columnwidth]{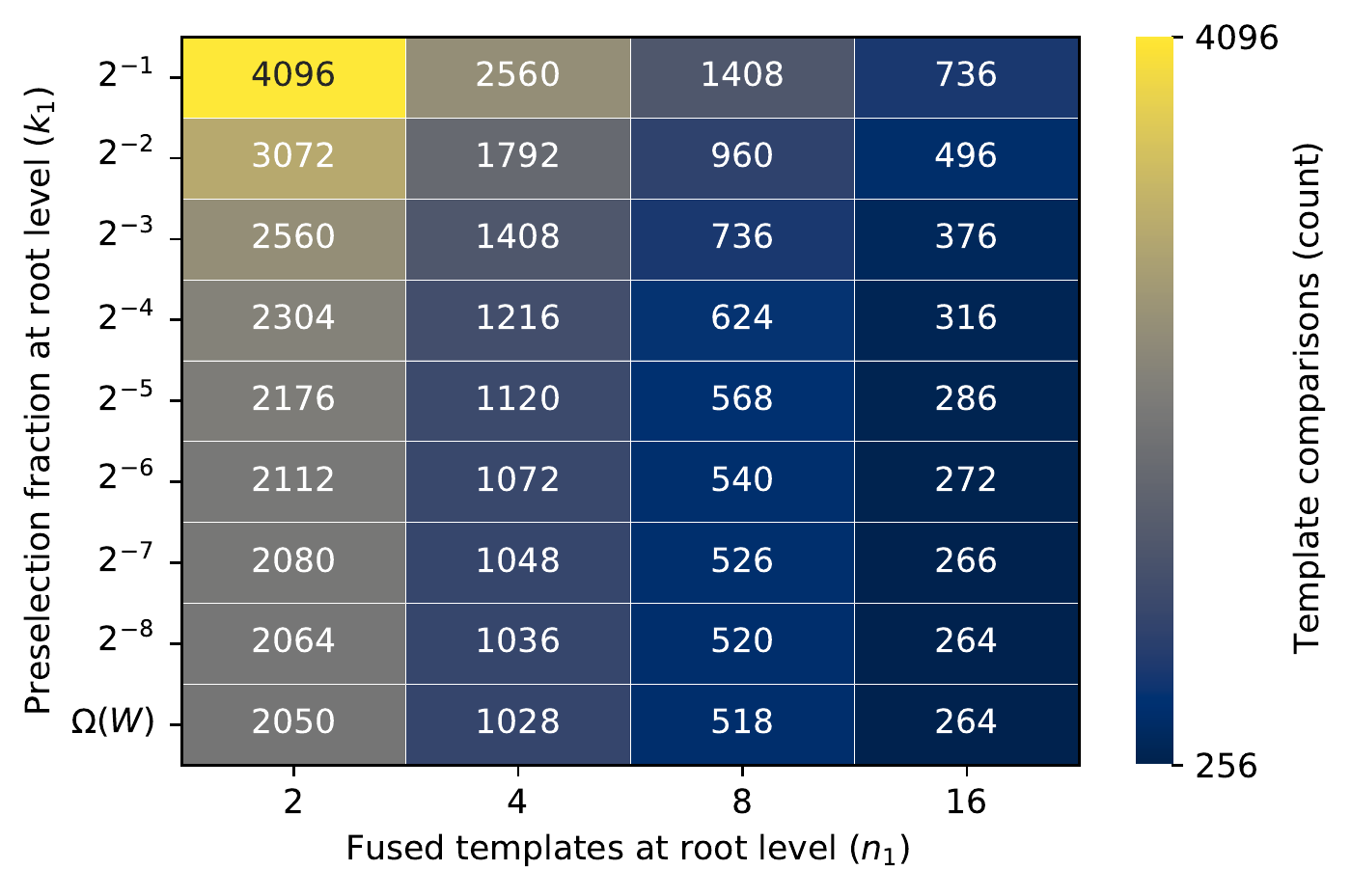}
\caption{Numbers of necessary template comparisons per identification transaction for $N = 4096$ using the proposed system}
\label{fig:workload_practical}
\end{figure}

Figure \ref{fig:workload_practical} shows the computational workload in terms of necessary template comparisons for the proposed indexing and retrieval method. In contrast to the general, theoretical overview from subsection \ref{subsec:proposedsystem_retrieval}, this figure pertains to the specific experimental setup described in section \ref{sec:experimentalsetup}. 

As noted in the theoretical analysis, the desired target area for the parameter ($n_{1}$ and $k_{1}$) selection lies in the bottom right corner of the matrix. For instance, given $n_{1} = 16$ and $k_{1} = 2^{-3}$, only 376 template comparisons are required for a biometric identification transaction, which is much lower than the 4096 template comparisons needed in the baseline scenario. In general, there exist several parameter configurations which result in the numbers of necessary template comparisons being significantly (between approximately 4 and 16 times) lower than those of a baseline retrieval method performing an exhaustive search.

In the next subsection, an analysis is conducted to determine whether the desirable configurations \wrt computational workload (\ie the rightmost part of the matrix in figure \ref{fig:workload_practical}) are also feasible \wrt biometric performance.

\subsection{Analysis of Pairing and Fusion Methods}
\label{subsec:results_configs}
From the point of view concentrated on biometric performance, the optimal selection of $n_{1}$ and $k_{1}$ parameters depends on following two factors:

\begin{enumerate}
\item The inherent discriminative power of the recognition system.
\item The information loss caused by the template fusion.
\end{enumerate} 

The information loss due to template fusion further depends on two factors: the used fusion method (section \ref{subsec:proposedsystem_fusion}) and the number of templates fused with each other (the $n_{1}$ parameter). As previously mentioned, the proposed indexing and retrieval scheme may cause false-negative errors when improperly configured, while the false-positive errors would remain unaffected or even slightly reduced through its application. To evaluate the two aforementioned factors, a closed-set identification scenario can be used and evaluated using CMC curves, which report the identification rate at given ranks (denoted $r$) in an ordered list of comparison scores between the probes and enrolment database.

\begin{figure}[!ht]
\centering
\subfloat[Random pairing]{\includegraphics[width=0.485\columnwidth]{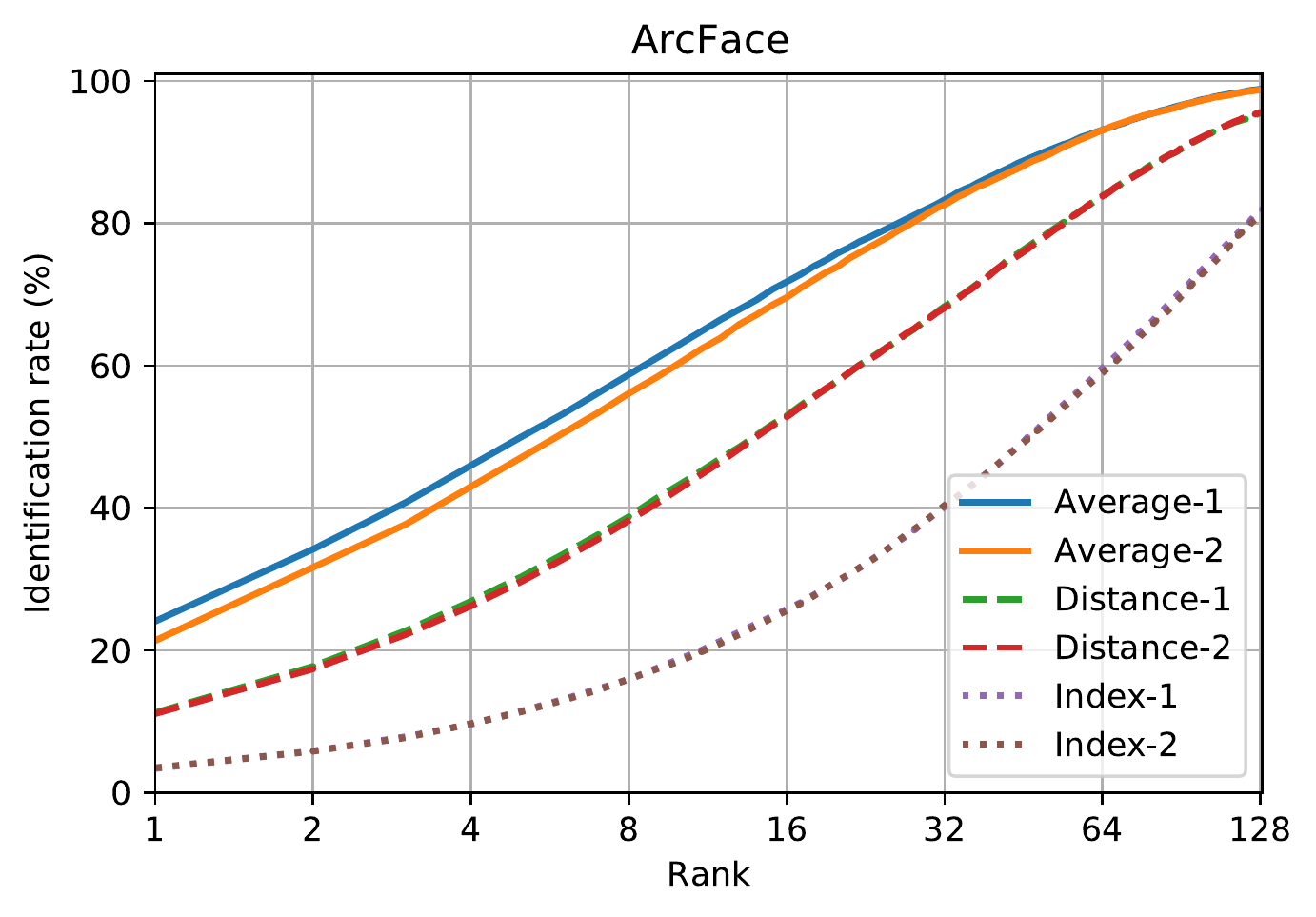} \hfill
\includegraphics[width=0.485\columnwidth]{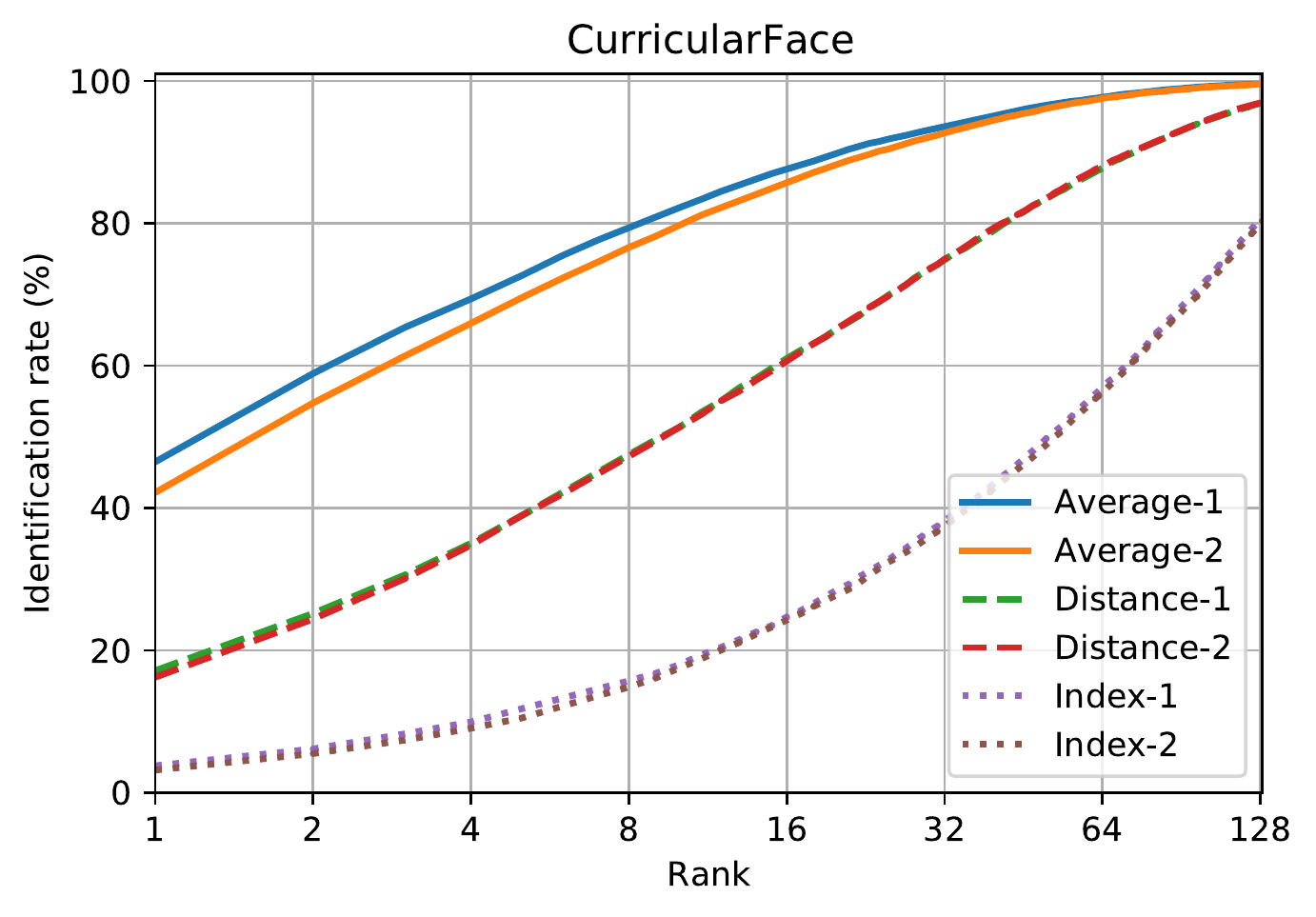}
} \\
\subfloat[Soft biometric pairing]{\includegraphics[width=0.485\columnwidth]{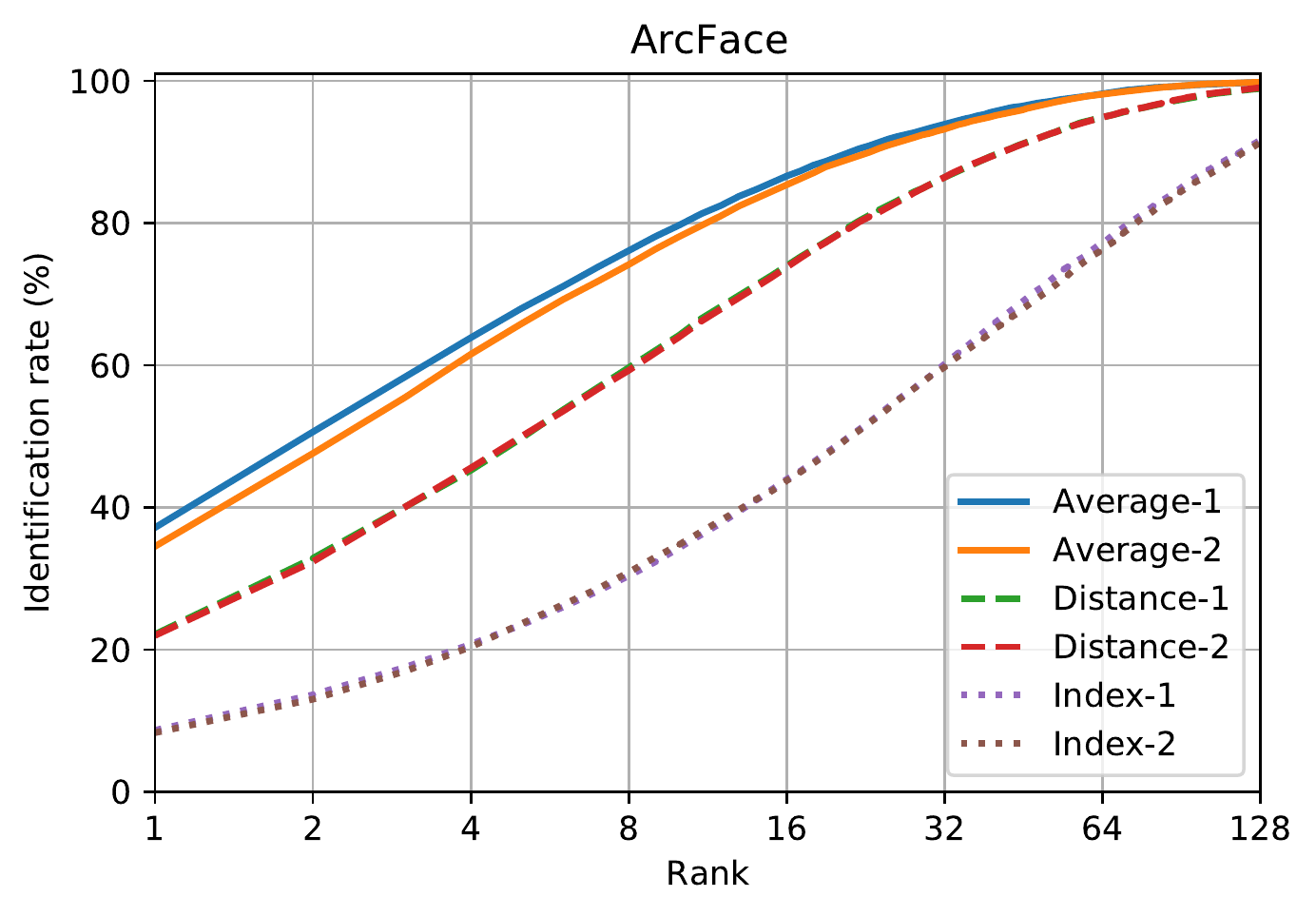} \hfill
\includegraphics[width=0.485\columnwidth]{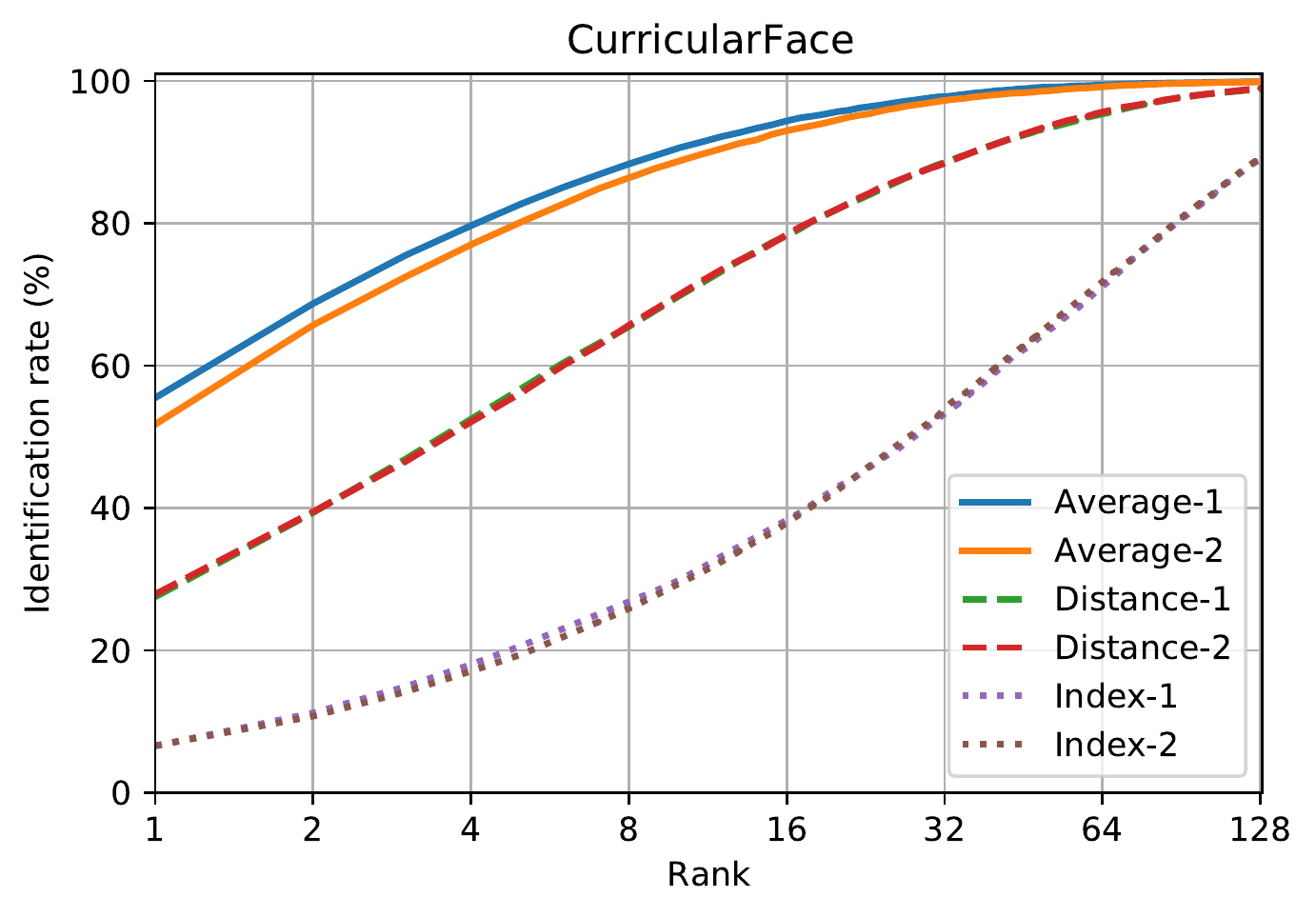}
} \\
\subfloat[Comparison score pairing]{\includegraphics[width=0.485\columnwidth]{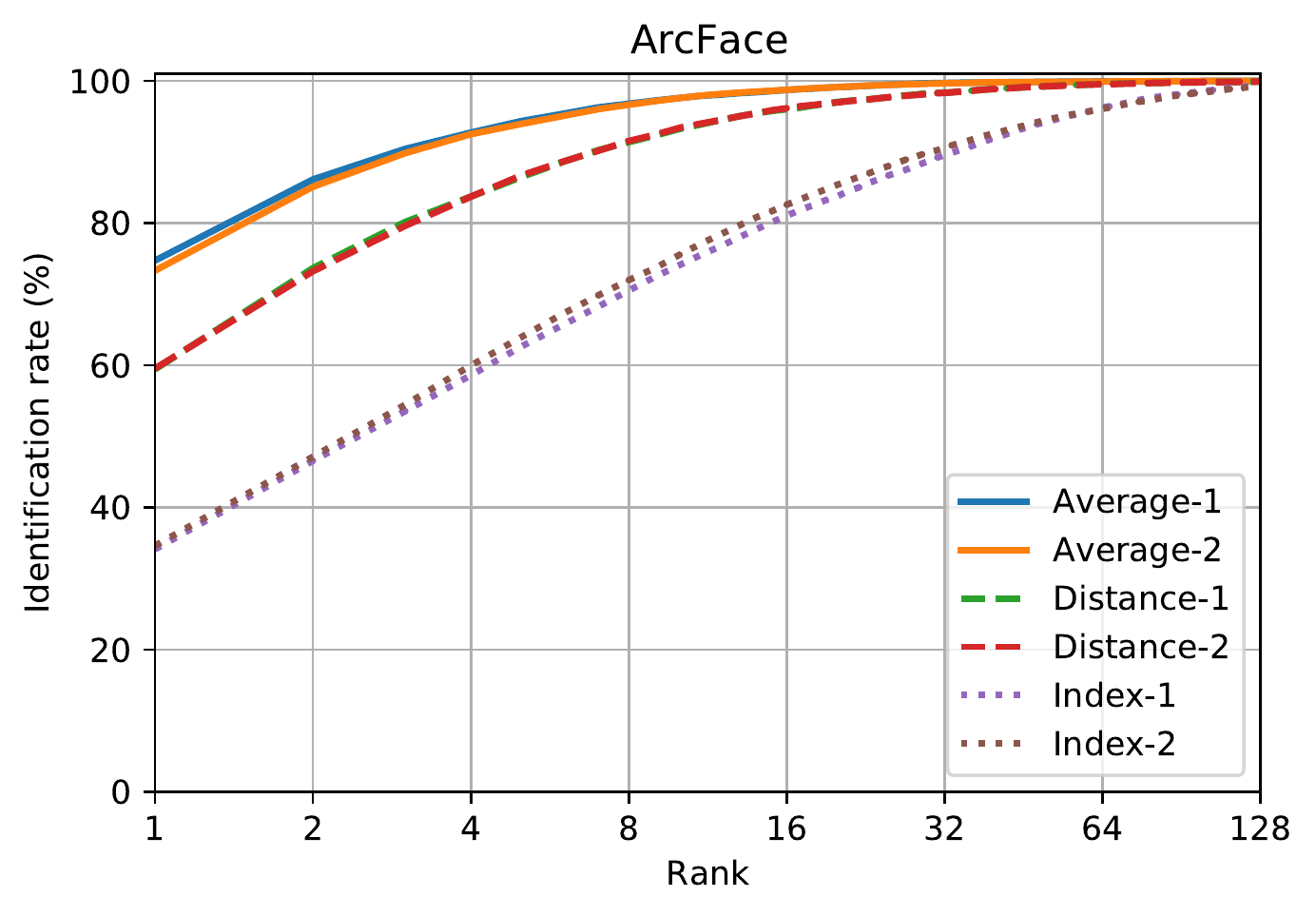} \hfill
\includegraphics[width=0.485\columnwidth]{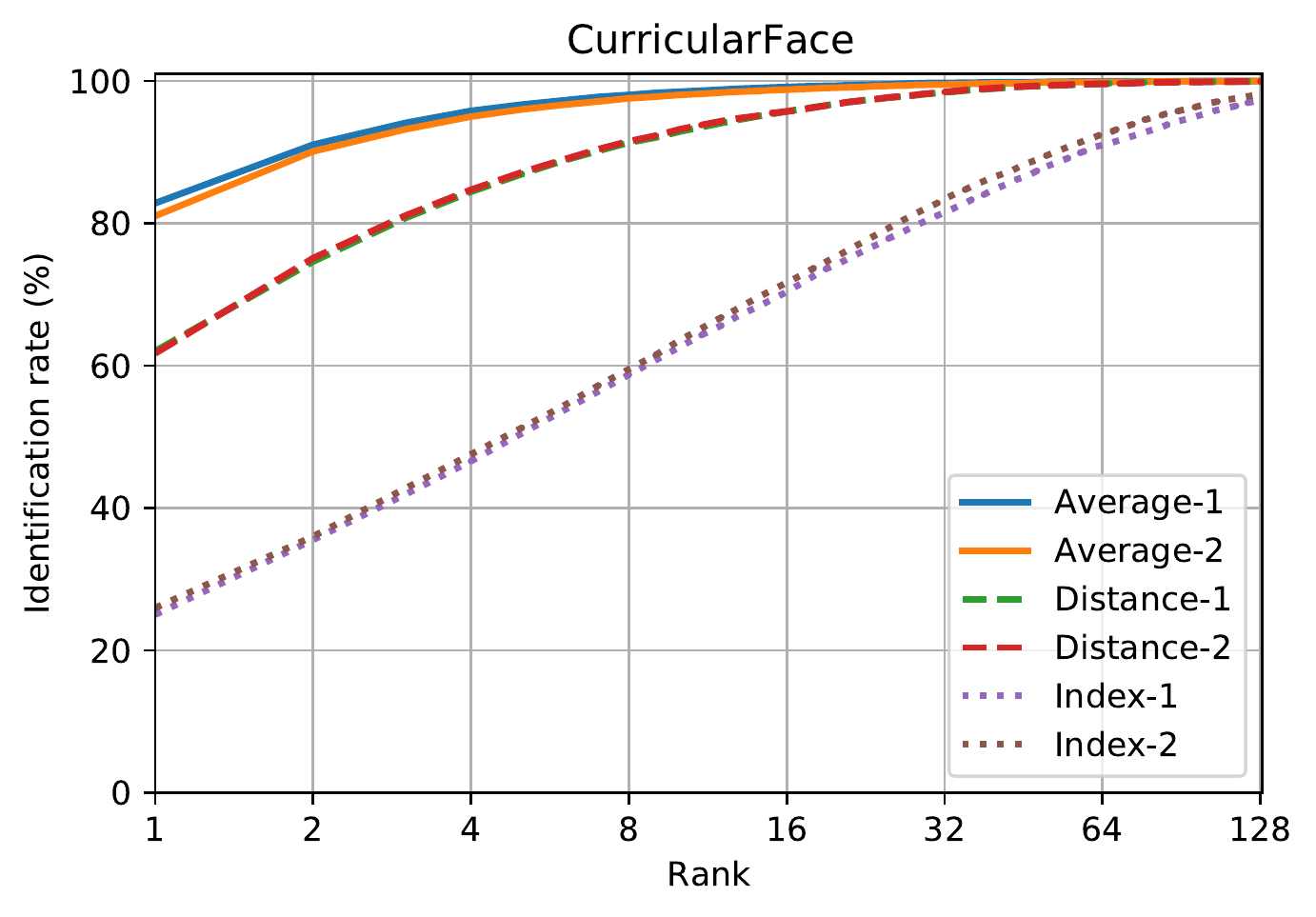}
} \\
\caption{CMC curves for experiment with $n_{1} = 16$ templates contributing to a fusion}
\label{fig:cmc_16}
\end{figure}

Figure \ref{fig:cmc_16} shows the CMC curves for the considered facial recognition systems, template pairing methods, and template fusion methods. Aiming at highest possible workload reduction (recall subsection \ref{subsec:results_workload}), $n_{1} = 16$ (\ie maximum rank is 256) is selected. Table \ref{table:identification_rates_16} shows the numeric values of identification rate for the specific ranks depicted on the x-axis in the figure.

\begin{table}[!ht]
\centering
\caption{Identification rates with $n_{1} = 16$ (in \%)}
\label{table:identification_rates_16}
\resizebox{\columnwidth}{!}{
\begin{tabular}{lllrrrrrrrr}
\toprule
\multirow{2}{*}{\textbf{Recognition}} & \multirow{2}{*}{\textbf{Pairing}} & \multirow{2}{*}{\textbf{Fusion}} & \multicolumn{8}{c}{\textbf{Rank}} \\ \cmidrule{4-11}
& & & \textbf{1} & \textbf{2} & \textbf{4} & \textbf{8} & \textbf{16} & \textbf{32} & \textbf{64} & \textbf{128} \\
\midrule
ArcFace & Random & Average-1 & 24.09 & 34.21 & 45.98 & 58.76 & 71.76 & 83.32 & 93.17 & 98.83 \\
& & Average-2 & 21.38 & 31.69 & 43.01 & 56.12 & 69.59 & 82.61 & 93.08 & 98.75 \\
& & Distance-1 & 11.24 & 17.76 & 26.89 & 38.77 & 52.97 & 68.34 & 83.90 & 95.52 \\
& & Distance-2 & 11.11 & 17.40 & 26.25 & 38.23 & 52.81 & 68.15 & 83.82 & 95.57 \\
& & Index-1 & 3.40 & 5.85 & 9.62 & 15.93 & 25.78 & 40.18 & 59.64 & 81.66 \\
& & Index-2 & 3.51 & 5.84 & 9.71 & 15.98 & 25.55 & 40.40 & 59.01 & 81.30 \\
\cmidrule{2-11}
& Soft biometric & Average-1 & 37.17 & 50.63 & 63.88 & 76.10 & 86.61 & 93.89 & 98.20 & 99.76 \\
& & Average-2 & 34.52 & 47.61 & 61.55 & 74.16 & 85.42 & 93.21 & 98.13 & 99.81 \\
& & Distance-1 & 22.15 & 32.86 & 45.23 & 59.64 & 73.98 & 86.43 & 94.85 & 98.95 \\
& & Distance-2 & 22.00 & 32.39 & 45.55 & 59.28 & 73.81 & 86.43 & 94.89 & 99.08 \\
& & Index-1 & 8.66 & 13.66 & 20.76 & 30.38 & 43.94 & 60.12 & 77.22 & 91.66 \\
& & Index-2 & 8.32 & 13.03 & 20.39 & 30.89 & 43.77 & 59.73 & 76.29 & 91.38 \\
\cmidrule{2-11}
& Comparison score & Average-1 & 74.72 & 86.13 & 92.72 & 96.78 & 98.72 & 99.68 & 99.91 & 99.99 \\
& & Average-2 & 73.31 & 85.11 & 92.50 & 96.64 & 98.73 & 99.66 & 99.90 & 99.99 \\
& & Distance-1 & 59.43 & 73.65 & 83.72 & 91.37 & 96.00 & 98.35 & 99.54 & 99.89 \\
& & Distance-2 & 59.54 & 73.22 & 83.71 & 91.54 & 96.14 & 98.31 & 99.51 & 99.90 \\
& & Index-1 & 34.12 & 46.56 & 58.59 & 70.48 & 81.00 & 89.56 & 96.23 & 99.31 \\
& & Index-2 & 34.61 & 47.10 & 59.96 & 71.98 & 82.58 & 90.59 & 96.06 & 99.25 \\
\midrule
CurricularFace & Random & Average-1 & 46.48 & 58.90 & 69.38 & 79.40 & 87.63 & 93.56 & 97.70 & 99.56 \\
& & Average-2 & 42.17 & 54.76 & 65.95 & 76.61 & 85.71 & 92.70 & 97.53 & 99.54 \\
& & Distance-1 & 17.15 & 25.19 & 35.05 & 47.51 & 61.01 & 74.87 & 87.77 & 96.84 \\
& & Distance-2 & 16.23 & 24.41 & 34.80 & 47.26 & 60.64 & 74.98 & 87.97 & 96.92 \\
& & Index-1 & 3.79 & 6.16 & 9.95 & 15.72 & 24.69 & 38.24 & 57.02 & 80.64 \\
& & Index-2 & 3.18 & 5.50 & 9.07 & 14.85 & 24.20 & 37.47 & 56.27 & 79.97 \\
\cmidrule{2-11}
& Soft biometric & Average-1 & 55.48 & 68.72 & 79.67 & 88.34 & 94.38 & 97.81 & 99.44 & 99.90 \\
& & Average-2 & 51.76 & 65.70 & 77.00 & 86.38 & 93.01 & 97.26 & 99.17 & 99.88 \\
& & Distance-1 & 27.52 & 39.36 & 52.48 & 65.48 & 78.26 & 88.54 & 95.37 & 98.92 \\
& & Distance-2 & 27.87 & 39.46 & 52.11 & 65.66 & 78.45 & 88.45 & 95.63 & 98.96 \\
& & Index-1 & 6.56 & 11.21 & 18.01 & 26.79 & 38.33 & 53.29 & 71.17 & 89.07 \\
& & Index-2 & 6.61 & 10.76 & 17.08 & 25.88 & 37.91 & 53.90 & 71.79 & 89.24 \\
\cmidrule{2-11}
& Comparison score & Average-1 & 82.83 & 91.04 & 95.80 & 98.04 & 99.14 & 99.72 & 99.93 & 99.99 \\
& & Average-2 & 81.03 & 90.10 & 94.99 & 97.55 & 98.77 & 99.50 & 99.86 & 99.98 \\
& & Distance-1 & 62.04 & 74.63 & 84.43 & 91.34 & 95.72 & 98.41 & 99.61 & 99.93 \\
& & Distance-2 & 61.73 & 75.10 & 84.71 & 91.53 & 95.72 & 98.45 & 99.59 & 99.94 \\
& & Index-1 & 25.03 & 35.54 & 46.59 & 58.73 & 70.37 & 81.54 & 90.99 & 97.39 \\
& & Index-2 & 25.97 & 36.02 & 47.48 & 59.50 & 71.67 & 83.45 & 92.50 & 98.12 \\
\bottomrule
\end{tabular}
}
\end{table}

Following observations can be made:
\begin{enumerate}
\item The proposed methods of intelligent pairing of subject templates to be fused result in a significant improvement of the identification rates \wrt to random pairings (\ie no optimisation).
\item The pairing of templates based on (non-mated) comparison scores performs much better than the pairing based on soft-biometric attributes of the data subjects.
\item The biometric performance across the three considered types of template fusion methods varies significantly. In all considered cases, the fusion methods based on averaging perform best, relatively closely followed by fusion methods based on distance from mean. The index-based fusion methods achieve a poor biometric performance. The differences between the fusion method variants within their respective method types are insignificant.
\item Although rank-1 identification rate is very low, both recognition systems quickly converge (for the best performing type of fusion methods) at 100\% well before the maximum rank of 256.
\item In general, CurricularFace performs slightly better than ArcFace. However, the differences are not very large and the general trends described above persist across both recognition systems. 
\end{enumerate}

Based on the above evaluations and observations, the selection of optimal configurations for pre-selection can be made. Accordingly, following choices are made:

\begin{LaTeXdescription}
\item[Template pairing] the method based on non-mated comparison scores is chosen.
\item[Template fusion] the method based on averaging the contributing templates is chosen.
\item[Number of fused templates] $n_{1} = 16$ can be used, as both recognition systems appear to exhibit sufficient discriminative power to compensate for the information loss caused by fusing so many templates.
\item[Fraction of preselected templates] To avoid too many pre-selection errors, configurations with $IR(r) > 99.5\%$ are considered. This condition is satisfied for both ArcFace and CurricularFace when $r \in \{32, 64, 128\}$, using the comparison score-based pairing and averaging-based fusion. These $r$ values correspond to $k_{1} \in \{2^{-1}, 2^{-2}, 2^{-3}\}$. For recognition systems with greater discriminative power, it is conceivable to achieve even lower $r$ and $k_{1}$ values, thereby facilitating higher workload reduction.
\end{LaTeXdescription}

\subsection{Overall Results}
\label{subsec:results_overall}
\begin{figure}[!ht]
\centering
\subfloat[ArcFace]{\includegraphics[width=\columnwidth]{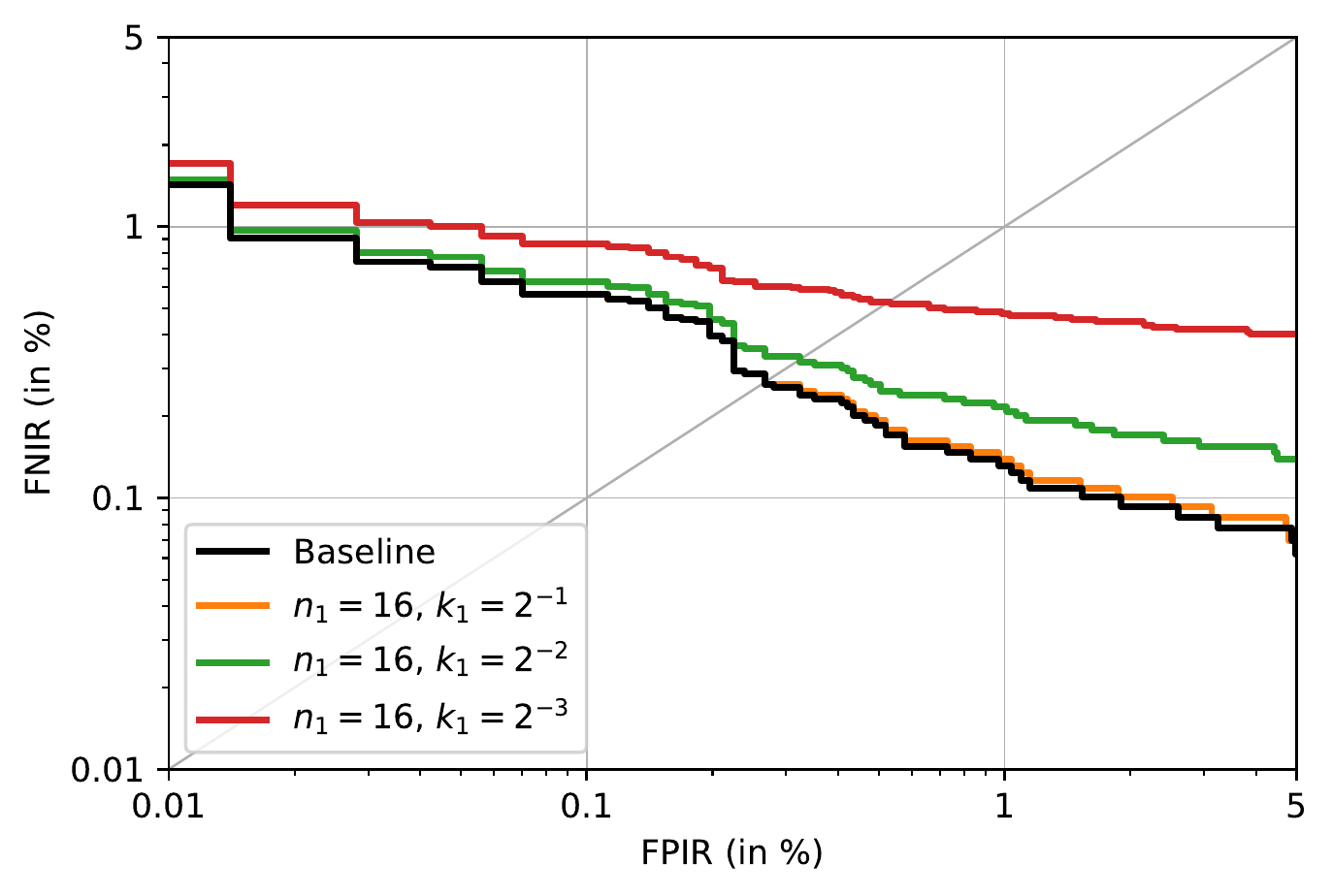}} \\
\subfloat[CurricularFace]{\includegraphics[width=\columnwidth]{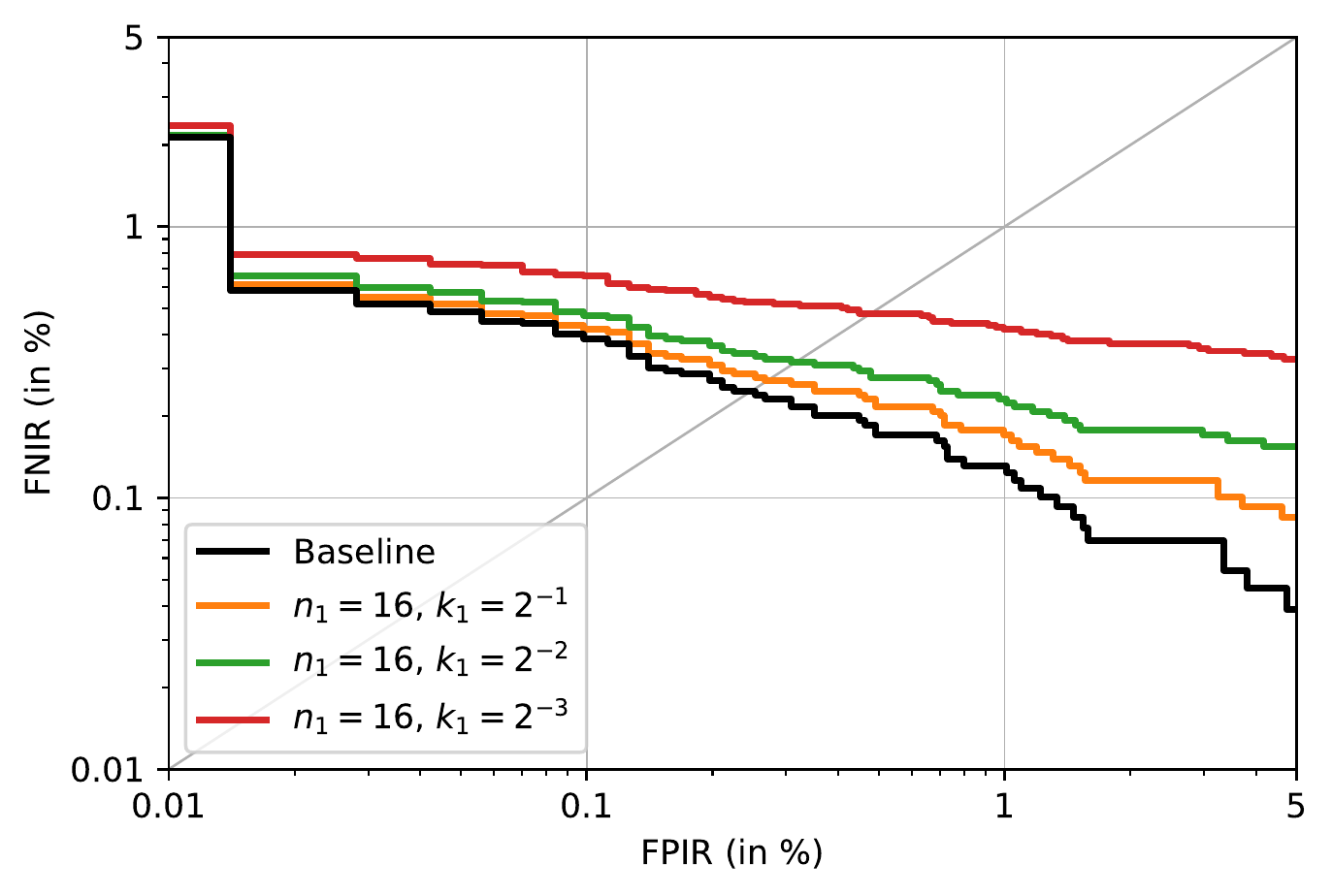}}
\caption{DET curves for the chosen parameter configurations}
\label{fig:det}
\end{figure}

To evaluate the overall performance of the proposed indexing and retrieval system, open-set and closed-set identification experiments are carried out for the configurations selected in subsections \ref{subsec:results_workload} and \ref{subsec:results_configs}. Figure \ref{fig:det} shows the obtained DET curves, while table \ref{table:results_summary} reports the numeric results using metrics described in subsection \ref{subsec:experimentalsetup_evaluationmetrics}.

\begin{table*}[!ht]
\centering
\caption{Summary of the proposed system results}
\label{table:results_summary}
\resizebox{0.75\textwidth}{!}{
\begin{tabular}{llrrrr}
\toprule
\multirow{2}{*}{\textbf{Recognition}} & \multirow{2}{*}{\textbf{Configuration}} & \multirow{2}{*}{\textbf{Workload}} & \multicolumn{2}{c}{\textbf{Open-set}} & \textbf{Closed-set} \\
\cmidrule(r){4-5} \cmidrule(l){6-6}
& & & \textbf{EER} & $\mathbf{FNIR_{1000}}$ & \textbf{RR-1} \\
\midrule
ArcFace & Baseline & 100.00\% & 0.27\% & 0.56\% & 99.96\% \\
& $n_{1} = 16$, $k_{1} = 2^{-1}$ & 17.97\% & 0.27\% & 0.56\% & 99.95\% \\
& $n_{1} = 16$, $k_{1} = 2^{-2}$ & 12.11\% & 0.32\% & 0.63\% & 99.87\% \\
& $n_{1} = 16$, $k_{1} = 2^{-3}$ & 9.18\% & 0.52\% & 0.87\% & 99.57\% \\
\midrule
CurricularFace & Baseline & 100.00\% & 0.25\% & 0.39\% & 99.98\% \\
& $n_{1} = 16$, $k_{1} = 2^{-1}$ & 17.97\% & 0.27\% & 0.42\% & 99.94\% \\
& $n_{1} = 16$, $k_{1} = 2^{-2}$ & 12.11\% & 0.32\% & 0.47\% & 99.86\% \\
& $n_{1} = 16$, $k_{1} = 2^{-3}$ & 9.18\% & 0.48\% & 0.66\% & 99.64\% \\
\bottomrule
\end{tabular}
}
\end{table*}

Following observations regarding computational workload and biometric performance can be made:

\begin{LaTeXdescription}
\item[ArcFace] All three chosen configurations perform similarly to the baseline. The most conservative one in terms of computational workload reduction, \ie $k_{1} = 2^{-1}$, achieves biometric performance essentially indistinguishable from that of the baseline, while simultaneously requiring only around 18\% of the computational workload that the baseline requires. The computational workload can be further reduced to less than 10\% of the baseline workload ($k_{1} = 2^{-3}$), while retaining a reasonable (albeit slightly reduced) biometric performance \wrt the baseline.
\item[CurricularFace] The results mirror those of ArcFace, thus indicating a generalisability of the proposed indexing and retrieval method. The achieved computational workload reduction is identical, as same parameter configurations have been used. The biometric performance of CurricularFace is slightly better than that of ArcFace, in particular at the $\text{FNIR}_{1000}$ operating point. The proposed system basically maintains the biometric performance of the baseline at $k_{1} \in \{2^{-1}, 2^{-2}\}$ for the practically relevant FPIR values, whereas $k_{1} = 2^{-3}$ yields an even lower computational workload at the cost of a slight reduction in biometric performance.
\end{LaTeXdescription}

In table \ref{table:results_executiontimes}, a summary of the computational requirements for the proposed protected indexing system is given in actual runtimes and storage usage for the off-the-shelf hardware mentioned in subsection \ref{subsec:experimentalsetup_homomorphicencryption}.

\begin{table}[!ht]
\centering
\caption{Approximate computational and data storage requirements using the hardware specified in subsection \ref{subsec:experimentalsetup_homomorphicencryption} for the proposed system with $N = 4096$ and 128 bits security}
\label{table:results_executiontimes}
\resizebox{\columnwidth}{!}{
\begin{tabular}{llrrr}
\toprule
\multirow{2}{*}{\textbf{Method}} & \multirow{2}{*}{\textbf{Configuration}} & \multicolumn{2}{c}{\textbf{Execution time}} & \multirow{2}{*}{\textbf{Storage}} \\ \cmidrule{3-4}
& & \textbf{DB Encryption} & \textbf{Identification} & \\
\midrule
CKKS & Baseline & {$\sim$}25 seconds & {$\sim$}4 hours & {$\sim$}2.2 GB \\ 
& $n_{1} = 16$, $k_{1} = 2^{-1}$ & {$\sim$}50 seconds & {$\sim$}40 minutes & {$\sim$}4.4 GB \\ 
& $n_{1} = 16$, $k_{1} = 2^{-2}$ & {$\sim$}50 seconds & {$\sim$}28 minutes & {$\sim$}4.4 GB \\ 
& $n_{1} = 16$, $k_{1} = 2^{-3}$ & {$\sim$}50 seconds & {$\sim$}21 minutes & {$\sim$}4.4 GB \\ 
\midrule
BFV & Baseline & {$\sim$}5 minutes & {$\sim$}42 minutes & {$\sim$}0.55 GB \\ 
& $n_{1} = 16$, $k_{1} = 2^{-1}$ & {$\sim$}10 minutes & {$\sim$}7.5 minutes & {$\sim$}1.1 GB \\ 
& $n_{1} = 16$, $k_{1} = 2^{-2}$ & {$\sim$}10 minutes & {$\sim$}5 minutes & {$\sim$}1.1 GB \\ 
& $n_{1} = 16$, $k_{1} = 2^{-3}$ & {$\sim$}10 minutes & {$\sim$}4 minutes & {$\sim$}1.1 GB \\ 
\midrule
NTRU & Baseline & {$\sim$}2 minutes & {$\sim$}1.5 minutes & {$\sim$}22.5 MB \\ 
& $n_{1} = 16$, $k_{1} = 2^{-1}$ & {$\sim$}4 minutes & {$\sim$}17 seconds & {$\sim$}45 MB \\ 
& $n_{1} = 16$, $k_{1} = 2^{-2}$ & {$\sim$}4 minutes & {$\sim$}11 seconds & {$\sim$}45 MB \\ 
& $n_{1} = 16$, $k_{1} = 2^{-3}$ & {$\sim$}4 minutes & {$\sim$}9 seconds & {$\sim$}45 MB \\ 
\bottomrule
\end{tabular}
}
\end{table}

It can be observed that:

\begin{itemize}
\item There exist massive differences in execution time and storage space usage between the benchmarked homomorphic encryption methods. The proposed indexing and retrieval method dramatically (order of magnitude) reduces the execution times of an identification transaction \wrt the baseline.
\item The one-time computational costs of encrypting the enrolment database and its index are negligible.
\item The execution times of the proposed system with BFV and especially CKKS based encryption do not suffice for real-time deployments, but could nevertheless be feasible whenever near-instantaneous system responses are not required.
\item Near-realtime runtimes and very low storage usage are achieved for the proposed system with NTRU-based encryption. This is mostly because the Hamming weight (\ie the sum of the differences between individual feature vector elements) cannot be computed in the encrypted domain using this scheme. On the other hand, in BFV- and CKKS-based schemes, the analogous sum can be (and is) computed in the encrypted domain. While in principle still secure and privacy-preserving \cite{Kolberg-NTRU-2019,Kolberg-FaceHE-BTP-BIOSIG-2020}, this means that using NTRU in the proposed system introduces a further trade-off between computational requirements and potential of some information leakage.
\end{itemize}

\subsection{Security Analysis}
\label{subsec:results_security}
The proposed system fulfils the biometric template protection objectives specified in ISO/IEC IS 24745 \cite{ISO11-TemplateProtection}:

\begin{LaTeXdescription}
\item[Unlinkability] a random factor is utilised in the encryption functions. Thus, encrypting an identical plaintext twice results in two different, unlinkable ciphertexts.
\item[Irreversibility] the used HE schemes are based on ideal lattices, \ie are post-quantum-secure \cite{bernstein2017post}. They provide encryption with the strength of 128, 192, or 256 bits\footnote{According to the \url{https://homomorphicencryption.org/} standard.}. There exists a trade-off between security and computational requirements -- as the encryption strength increases, so does the computational complexity.
\item[Renewability] the HE key pair can be exchanged, whereupon the biometric templates in the enrolment database and index can be re-encrypted.
\item[Performance preservation] the comparator used in the homomorphically encrypted domain is functionally identical to that of the plaintext domain, \ie it yields the same comparison scores. The biometric performance of the proposed indexing system is nearly identical to that of the baseline. 
\end{LaTeXdescription}

While traditional encryption schemes may provide stronger security guarantees than homomorphic encryption, this does not constitute the actual limiting factor \wrt facial biometrics. The entropy of facial embeddings is considered to be much lower than the aforementioned achievable cryptographic protection levels. For example, \eg in \cite{gong2019intrinsic}, it has been shown that while typical facial embeddings extracted by deep neural networks consist of 512 values, their intrinsic dimensionality is much lower (more than an order of magnitude). In other words, it is more feasible (albeit still extremely difficult) to guess a sufficiently similar facial biometric template than to guess the encryption keys. 

Finally, note that such attacks aimed at guessing the biometric templates and/or encryption keys (or other secrets) are not limited to the applications of homomorphic encryption for the purpose of biometric template protection. Other types of dedicated biometric template protection approaches (recall subsection \ref{subsec:background_templateprotection}) as well as classic general-purpose (non-homomorphic) encryption must likewise address those challenges.

\subsection{Scalability}
\label{subsec:results_scalability}
As the size of the enrolment database increases, following factors within the proposed system need to be considered:

\begin{LaTeXdescription}
\item[Pairing] although the pairing algorithm is computationally intensive, its computational costs could be easily mitigated by distributing the computations or additionally binning the enrolment database. It should also be noted that an increased size of the enrolment database would result in a larger probability of finding suitable pairings -- especially for the outlier subjects (and hence an increased discriminative power of the system). 
\item[Fusion] the operations for fusing the templates are implemented efficiently using vectorised operations; these computational costs are generally are negligible, \eg in comparison with those required for template pairing. Furthermore, this part of the proposed system's pipeline can be trivially parallelised and/or distributed.
\item[Encryption] the computational costs of encrypting the enrolment database and index are generally very low and can additionally be trivially parallelised. The amount of RAM required to pre-load (for use during retrieval) the entire enrolment database and its index is approximately twice that of the baseline.
\item[Retrieval] the computational workload of the proposed system scales \textit{sub-linearly} \wrt to the number of enrolled subjects (as opposed to a typical baseline, which typically scales linearly). Due to a flexible design of the proposed system, a dynamic adjustment (\wrt enrolment database size) of the decision thresholds and pre-selection subset sizes is possible. Lastly, the underlying concepts in the proposed indexing and retrieval system can be trivially distributed or parallelised.  
\end{LaTeXdescription}

The pairing, fusion, and encryption operations are computed infrequently and offline; they thus do not directly influence the (online) retrieval time. Considering the execution timings in table \ref{table:results_executiontimes}, it is important to note that the experiments were carried out in a single-threaded environment on an ordinary laptop. Taking advantage of parallelisation or distribution of the computations, as well as utilising more powerful hardware, these execution times could be vastly lowered (\cf \cite{Engelsma-HomomorphicIdentification-2020}).

\section{Conclusion}
\label{sec:conclusion}
In this article, a method of computationally efficient indexing and retrieval of biometric data has been presented. The proposed indexing method relies on intelligent pairing of facial parent templates based on their similarity (in terms of soft biometrics or non-mated comparison scores), followed by feature-level fusion. The created search structure facilitates a multi-step biometric identification retrieval, whereby the retrieved candidate lists are successively shortened in each step of the cascade.

In a comprehensive experimental evaluation, several different pairing and fusion methods were benchmarked for the indexing step using two modern, open-source face recognition systems. Using standardised evaluation protocols and metrics, the proposed method was shown to achieve a biometric performance nearly identical to that of an exhaustive search-based baseline; simultaneously the computational workload of biometric identification transactions has been substantially reduced (down to ${\sim}10\%$). In other words, by using the proposed system during biometric identification, a tenfold reduction in the required computational effort is possible with no negative impact on the biometric performance. By integrating homomorphic encryption, the proposed system achieves post-quantum-security and the biometric template protection objectives of unlinkability, irreversibility, and renewability.

In summary, the proposed system achieves a very good balance between biometric performance, computational efficiency, and privacy protection for biometric identification scenarios.

\section*{Acknowledgements}
\label{sec:acknowledgements}
This research work has been funded by the German Federal Ministry of Education and Research and the Hessian Ministry of Higher Education, Research, Science and the Arts within their joint support of the National Research Center for Applied Cybersecurity ATHENE and the European Union’s Horizon 2020  research and innovation programme under the Marie Skłodowska-Curie grant agreement No. 860813 - TReSPAsS-ETN.

\bibliographystyle{IEEEtran}
\bibliography{references}

\begin{thebibliography}{100}
\providecommand{\url}[1]{#1}
\csname url@samestyle\endcsname
\providecommand{\newblock}{\relax}
\providecommand{\bibinfo}[2]{#2}
\providecommand{\BIBentrySTDinterwordspacing}{\spaceskip=0pt\relax}
\providecommand{\BIBentryALTinterwordstretchfactor}{4}
\providecommand{\BIBentryALTinterwordspacing}{\spaceskip=\fontdimen2\font plus
\BIBentryALTinterwordstretchfactor\fontdimen3\font minus
  \fontdimen4\font\relax}
\providecommand{\BIBforeignlanguage}[2]{{%
\expandafter\ifx\csname l@#1\endcsname\relax
\typeout{** WARNING: IEEEtran.bst: No hyphenation pattern has been}%
\typeout{** loaded for the language `#1'. Using the pattern for}%
\typeout{** the default language instead.}%
\else
\language=\csname l@#1\endcsname
\fi
#2}}
\providecommand{\BIBdecl}{\relax}
\BIBdecl

\bibitem{Das-MobileBiometrics-2018}
A.~Das, C.~Galdi, H.~Han, R.~Ramachandra, J.-L. Dugelay, and A.~Dantcheva,
  ``Recent advances in biometric technology for mobile devices,'' in
  \emph{International Conference on Biometrics Theory, Applications and Systems
  ({BTAS})}.\hskip 1em plus 0.5em minus 0.4em\relax IEEE, October 2018, pp.
  1--11.

\bibitem{EULisa-EURODAC-2016}
S.~European Union Agency for the Operational Management of Large-Scale IT
  Systems in the Area~of Freedom and Justice, ``Eurodac storage capacity
  increased,''
  \url{https://www.eulisa.europa.eu/Newsroom/News/Pages/Eurodac-storage-capacity-increased.aspx},
  April 2016, last accessed: \today.

\bibitem{SmartBorders-EU-2018}
{European Commission}, ``Smart borders,''
  \url{https://ec.europa.eu/home-affairs/what-we-do/policies/borders-and-visas/smart-borders_en},
  2018, last accessed: \today.

\bibitem{Thales-IDENT-2021}
Thales, ``{DHS}'s automated biometric identification system {IDENT} - the heart
  of biometric visitor identification in the {USA},''
  \url{https://www.thalesgroup.com/en/markets/digital-identity-and-security/government/customer-cases/ident-automated-biometric-identification-system},
  January 2021, last accessed: \today.

\bibitem{Moses-AFIS-2010}
K.~R. Moses, P.~Higgins, M.~McCabe, S.~Probhakar, and S.~Swann,
  \emph{Fingerprint Sourcebook}.\hskip 1em plus 0.5em minus 0.4em\relax US
  Department of Justice, 2010, ch. Automated Fingerprint Identification System
  ({AFIS}), pp. 1--33.

\bibitem{FBI-CODIS-2021}
F.~B. of~Investigation, ``{CODIS} - {NDIS} statistics,''
  \url{https://www.fbi.gov/services/laboratory/biometric-analysis/codis/ndis-statistics},
  March 2021, last accessed: \today.

\bibitem{Thales-AFIS-2021}
Thales, ``{A}utomated {F}ingerprint {I}dentification {S}ystem ({AFIS}) - a
  short history,''
  \url{https://www.thalesgroup.com/en/markets/digital-identity-and-security/government/biometrics/afis-history},
  April 2021, last accessed: \today.

\bibitem{UIDAI-Aadhaar-2012}
{Unique Identification Authority of India}, ``Role of biometric technology in
  {A}adhaar enrollment,'' {UIDAI}, Tech. Rep., January 2012.

\bibitem{Dalwai-Aadhaar-2014}
A.~Dalwai, ``Aadhaar technology and architecture: principles, design, best
  practices and key lessons,'' Unique Identification Authority of India
  (UIDAI), Tech. Rep., March 2014.

\bibitem{Bowyer-IrisElection-2015}
K.~W. Bowyer, E.~Ortiz, and A.~Sgroi, ``Iris recognition technology evaluated
  for voter registration in {S}omaliland,'' \emph{Biometric Technology Today},
  vol. 2015, no.~2, pp. 5--8, February 2015.

\bibitem{CEPPS-Congo-2018}
C.~for Elections and P.~P. Strengthening, ``Assessment of electoral
  preparations in the {Democratic Republic of the Congo},'' CEPPS, Tech. Rep.,
  May 2018.

\bibitem{UIDAI-Dashboard}
{Unique Identification Authority of India}, ``Aadhaar dashboard,''
  \url{https://www.uidai.gov.in/aadhaar_dashboard/}, 2021, last accessed:
  \today.

\bibitem{Pascu-BiometricMarketValue-2020}
L.~Pascu, ``Global biometrics market to surpass \$45{B} by 2024, reports
  {F}rost \& {S}ullivan,''
  \url{https://www.biometricupdate.com/202003/global-biometrics-market-to-surpass-45b-by-2024-reports-frost-sullivan},
  March 2020.

\bibitem{Drozdowski-WorkloadSurvey-IET-2019}
P.~Drozdowski, C.~Rathgeb, and C.~Busch, ``Computational workload in biometric
  identification systems: An overview,'' \emph{IET Biometrics}, vol.~8, no.~6,
  pp. 351--368, November 2019.

\bibitem{FRVT-Identification-2020}
N.~I. of~Standards and Technology, ``{FRVT 1:N} identification,''
  \url{https://pages.nist.gov/frvt/html/frvt1N.html}, April 2021.

\bibitem{IREX-2018}
N.~I. of~Standards and T.~(NIST), ``Iris exchange ({IREX}),''
  \url{https://www.nist.gov/programs-projects/iris-exchange-irex-overview},
  2018, last accessed: \today.

\bibitem{Bust-DHSRally-2019}
C.~Burt, ``{DHS S\&T} biometric technology rally results suggest face best for
  fast processing,''
  \url{https://www.biometricupdate.com/201909/dhs-st-biometric-technology-rally-results-suggest-face-best-for-fast-processing},
  September 2019, last accessed: \today.

\bibitem{EU-GDPR-2016}
{European Parliament}, ``Regulation (eu) 2016/679,'' \emph{Official Journal of
  the {European Union}}, vol. L119, pp. 1--88, April 2016.

\bibitem{5396650}
A.~Cavoukian and A.~Stoianov, \emph{Biometric Encryption: The New Breed of
  Untraceable Biometrics}.\hskip 1em plus 0.5em minus 0.4em\relax Wiley-IEEE
  Press, 2010, pp. 655--718.

\bibitem{Rathgeb-TemplateProtection-EURASIP-2011}
C.~Rathgeb and A.~Uhl, ``A survey on biometric cryptosystems and cancelable
  biometrics,'' \emph{{EURASIP} Journal on Information Security}, 2011.

\bibitem{Nandakumar-TemplateProtection-IEEE-2015}
K.~Nandakumar and A.~K. Jain, ``Biometric template protection: Bridging the
  performance gap between theory and practice,'' \emph{Signal Processing
  Magazine}, vol.~32, no.~5, pp. 88--100, September 2015.

\bibitem{ISO11-TemplateProtection}
I.~J. S. . I.~S. techniques, \emph{{ISO/IEC} 24745:2011. Information technology
  -- Security techniques -- Biometric information protection}, International
  Organization for Standardization and International Electrotechnical
  Committee, June 2011.

\bibitem{Dong-face-identification-index-2020}
X.~Dong, S.~Kim, Z.~Jin, J.~Y. Hwang, S.~Cho, and A.~B.~J. Teoh, ``Open-set
  face identification with index-of-max hashing by learning,'' \emph{Pattern
  Recognition}, vol. 103, p. 107277, 2020.

\bibitem{Murakami-CancelableIndexing-2019}
T.~Murakami, R.~Fujita, T.~Ohki, Y.~Kaga, M.~Fujio, and K.~Takahashi,
  ``Cancelable permutation-based indexing for secure and efficient biometric
  identification,'' \emph{IEEE Access}, vol.~7, pp. 45\,563--45\,582, April
  2019.

\bibitem{Sardar-novel-cancelable-face-hashing-2020}
A.~Sardar, S.~Umer, C.~Pero, and M.~Nappi, ``A novel cancelable {FaceHashing}
  technique based on non-invertible transformation with encryption and
  decryption template,'' \emph{IEEE Access}, vol.~8, pp. 105\,263--105\,277,
  2020.

\bibitem{Boddeti-HE-2019}
V.~N. Boddeti, ``Secure face matching using fully homomorphic encryption,'' in
  \emph{International Conference on Biometrics Theory, Applications and Systems
  ({BTAS})}, 2019, pp. 1--10.

\bibitem{Drozdowski-HomomorphicIdentificationFace-BIOSIG-2019}
P.~Drozdowski, N.~Buchmann, C.~Rathgeb, M.~Margraf, and C.~Busch, ``On the
  application of homomorphic encryption to face identification,'' in
  \emph{International Conference of the Biometrics Special Interest Group
  ({BIOSIG})}, September 2019, pp. 1--8.

\bibitem{Kolberg-FaceHE-BTP-BIOSIG-2020}
J.~Kolberg, P.~Drozdowski, M.~Gomez-Barrero, C.~Rathgeb, and C.~Busch,
  ``Efficiency analysis of post-quantum-secure face template protection schemes
  based on homomorphic encryption,'' in \emph{International Conference of the
  Biometrics Special Interest Group ({BIOSIG})}, 2020, pp. 1--4.

\bibitem{Engelsma-HomomorphicIdentification-2020}
J.~J. Engelsma, A.~K. Jain, and V.~N. Boddeti, ``{HERS}: Homomorphically
  encrypted representation search,'' \emph{arXiv preprint arXiv:2003.12197},
  2020.

\bibitem{ISO-PerformanceReporting-2021}
{ISO/IEC JTC1 SC37 Biometrics}, \emph{{ISO/IEC} 19795-1:2021. Information
  Technology -- Biometric Performance Testing and Reporting -- Part~1:
  Principles and Framework}, International Organization for Standardization and
  International Electrotechnical Committee, 2021.

\bibitem{Ross-HandbookMultibiometrics-Springer-2006}
A.~Ross, K.~Nandakumar, and A.~K. Jain, \emph{Handbook of
  multibiometrics}.\hskip 1em plus 0.5em minus 0.4em\relax Springer, 2006.

\bibitem{ISO-Fusion-2015}
{ISO/IEC JTC1 SC37 Biometrics}, \emph{ISO/IEC TR 24722:2015. Information
  technology -- Biometrics -- Multimodal and other multibiometric fusion},
  2nd~ed., December 2015.

\bibitem{Jain-FingerprintMosaicking-2002}
A.~Jain and A.~Ross, ``Fingerprint mosaicking,'' in \emph{International
  Conference on Acoustics, Speech, and Signal Processing ({ICASSP})},
  vol.~4.\hskip 1em plus 0.5em minus 0.4em\relax IEEE, May 2002, pp.
  IV--4064--IV--4067.

\bibitem{Kusuma-PCAFusion-2011}
G.~P. Kusuma and C.-S. Chua, ``{PCA}-based image recombination for multimodal
  {2D + 3D} face recognition,'' \emph{Image and Vision Computing}, vol.~29,
  no.~5, pp. 306--316, April 2011.

\bibitem{Kanhangad-Fusion-2011}
V.~Kanhangad, A.~Kumar, and D.~Zhang, ``Contactless and pose invariant
  biometric identification using hand surface,'' \emph{Transactions on Image
  Processing ({TIP})}, vol.~20, no.~5, pp. 1415--1424, May 2011.

\bibitem{Yan-Fusion-2015}
X.~Yan, W.~Kang, F.~Deng, and Q.~Wu, ``Palm vein recognition based on
  multi-sampling and feature-level fusion,'' \emph{Neurocomputing}, vol. 151,
  no.~2, pp. 798--807, March 2015.

\bibitem{Snelick-Fusion-2003}
R.~Snelick, M.~Indovina, J.~Yen, and A.~Mink, ``Multimodal biometrics: issues
  in design and testing,'' in \emph{International Conference on Multimodal
  Interfaces}.\hskip 1em plus 0.5em minus 0.4em\relax ACM, November 2003, pp.
  68--72.

\bibitem{Jain-ScoreNormalization-2005}
A.~K. Jain, K.~Nandakumar, and A.~Ross, ``Score normalization in multimodal
  biometric systems,'' \emph{Pattern recognition}, vol.~38, no.~12, pp.
  2270--2285, December 2005.

\bibitem{Abaza-Fusion-2009}
A.~Abaza and A.~Ross, ``Quality based rank-level fusion in multibiometric
  systems,'' in \emph{International Conference on Biometrics: Theory,
  Applications, and Systems ({BTAS})}.\hskip 1em plus 0.5em minus 0.4em\relax
  IEEE, September 2009, pp. 1--6.

\bibitem{Kumar-Fusion-2011}
A.~Kumar and S.~Shekhar, ``Personal identification using multibiometrics
  rank-level fusion,'' \emph{Transactions on Systems, Man, and Cybernetics,
  Part C (Applications and Reviews)}, vol.~41, no.~5, pp. 743--752, September
  2011.

\bibitem{Prabhakar-Fusion-2002}
S.~Prabhakar and A.~K. Jain, ``Decision-level fusion in fingerprint
  verification,'' \emph{Pattern Recognition}, vol.~35, no.~4, pp. 861--874,
  April 2002.

\bibitem{Paul-Fusion-2014}
P.~P. Paul, M.~L. Gavrilova, and R.~Alhajj, ``Decision fusion for multimodal
  biometrics using social network analysis,'' \emph{Transactions on Systems,
  Man, and Cybernetics: Systems}, vol.~44, no.~11, pp. 1522--1533, November
  2014.

\bibitem{Dinca-FusionSurvey-2017}
L.~M. Dinca and G.~P. Hancke, ``The fall of one, the rise of many: A survey on
  multi-biometric fusion methods,'' \emph{IEEE Access}, vol.~5, pp. 6247--6289,
  April 2017.

\bibitem{Singh-Fusion-2019}
M.~Singh, R.~Singh, and A.~Ross, ``A comprehensive overview of biometric
  fusion,'' \emph{Information Fusion}, vol.~52, pp. 187--205, December 2019.

\bibitem{Daugman-DecisionLandscapes-UCAM-2000}
J.~Daugman, ``Biometric decision landscapes,'' University of Cambridge -
  Computer Laboratory, Tech. Rep. UCAM-CL-TR-482, January 2000.

\bibitem{Kavati-SearchReductionSurvey-IGI-2018}
I.~Kavati, M.~Prasad, and C.~Bhagvati, ``Search space reduction in biometric
  databases: a review,'' in \emph{Computer Vision: Concepts, Methodologies,
  Tools, and Applications}.\hskip 1em plus 0.5em minus 0.4em\relax IGI Global,
  2018, ch.~11, pp. 1600--1626.

\bibitem{Drozdowski-DeepFaceBinarisation-ICIP-2018}
P.~Drozdowski, F.~Struck, C.~Rathgeb, and C.~Busch, ``Benchmarking binarisation
  schemes for deep face templates,'' in \emph{International Conference on Image
  Processing ({ICIP})}.\hskip 1em plus 0.5em minus 0.4em\relax IEEE, October
  2018, pp. 191--195.

\bibitem{Gehrmann-MetadataFiltering-2019}
C.~Gehrmann, M.~Rodan, and N.~J{\"o}nsson, ``Metadata filtering for
  user-friendly centralized biometric authentication,'' \emph{EURASIP Journal
  on Information Security}, vol. 2019, no.~1, p.~7, June 2019.

\bibitem{Dantcheva-SoftBiometricsSurvey-TIFS-2016}
A.~Dantcheva, P.~Elia, and A.~Ross, ``What else does your biometric data
  reveal? {A} survey on soft biometrics,'' \emph{Transactions on Information
  Forensics and Security ({TIFS})}, vol.~11, no.~3, pp. 441--467, March 2016.

\bibitem{Gentile-TwoStageIris-BTAS-2009}
J.~E. Gentile, N.~Ratha, and J.~Connell, ``An efficient, two-stage iris
  recognition system,'' in \emph{International Conference on Biometrics:
  Theory, Applications and Systems ({BTAS})}.\hskip 1em plus 0.5em minus
  0.4em\relax IEEE, September 2009, pp. 211--215.

\bibitem{Billeb-SpeakerTwoStage-BIOSIG-2014}
S.~Billeb, C.~Rathgeb, M.~Buschbeck, H.~Reininger, and K.~Kasper, ``Efficient
  two-stage speaker identification based on universal background models,'' in
  \emph{International Conference of the Biometrics Special Interest Group
  ({BIOSIG})}.\hskip 1em plus 0.5em minus 0.4em\relax IEEE, September 2014, pp.
  1--6.

\bibitem{Pflug-HistogramBinarisation-CYBCONF-2015}
A.~Pflug, C.~Rathgeb, U.~Scherhag, and C.~Busch, ``Binarization of spectral
  histogram models: An application to efficient biometric identification,'' in
  \emph{International Conference on Cybernetics ({CYBCONF})}.\hskip 1em plus
  0.5em minus 0.4em\relax IEEE, June 2015, pp. 501--506.

\bibitem{Drozdowski-BloomFilterIndexing-IET-2018}
P.~Drozdowski, C.~Rathgeb, and C.~Busch, ``{B}loom filter-based search
  structures for indexing and retrieving {Iris-Codes},'' \emph{IET Biometrics},
  vol.~7, no.~3, pp. 260--268, May 2018.

\bibitem{Iloanusi-FingerprintIndexing-PRL-2014}
O.~N. Iloanusi, ``Fusion of finger types for fingerprint indexing using
  minutiae quadruplets,'' \emph{Pattern Recognition Letters}, vol.~38, pp.
  8--14, 2014.

\bibitem{Drozdowski-MultiIrisIndexing-IJCB-2017}
P.~Drozdowski, C.~Rathgeb, and C.~Busch, ``Multi-iris indexing and retrieval:
  Fusion strategies for {B}loom filter-based search structures,'' in
  \emph{International Joint Conference on Biometrics ({IJCB})}.\hskip 1em plus
  0.5em minus 0.4em\relax IEEE, October 2017, pp. 46--53.

\bibitem{Jayaraman-MultimodalIndexing-ICISS-2008}
U.~Jayaraman, S.~Prakash, and P.~Gupta, ``Indexing multimodal biometric
  databases using kd-tree with feature level fusion,'' in \emph{International
  Conference on Information Systems Security}.\hskip 1em plus 0.5em minus
  0.4em\relax Springer, 2008, pp. 221--234.

\bibitem{Gyaourova-IndexCodes-CVPRW-2009}
A.~Gyaourova and A.~Ross, ``A coding scheme for indexing multimodal biometric
  databases,'' in \emph{Conference on Computer Vision and Pattern Recognition
  Workshops ({CVPRW})}.\hskip 1em plus 0.5em minus 0.4em\relax IEEE, June 2009,
  pp. 93--98.

\bibitem{Gyaourova-IndexCodes-TIFS-2012}
A.~Gyaourova and A.~Ross, ``Index codes for multibiometric pattern retrieval,''
  \emph{Transactions on Information Forensics and Security ({TIFS})}, vol.~7,
  no.~2, pp. 518--529, April 2012.

\bibitem{Drozdowski-Kstage-2019}
P.~Drozdowski, C.~Rathgeb, B.-A. Mokro{\ss}, and C.~Busch, ``Multi-biometric
  identification with cascading database filtering,'' \emph{Transactions on
  Biometrics, Behavior, and Identity Science ({TBIOM})}, vol.~2, no.~3, pp.
  210--222, July 2020.

\bibitem{Drozdowski-MorphingPreselection-ICASSP-2019}
P.~Drozdowski, C.~Rathgeb, and C.~Busch, ``Turning a vulnerability into an
  asset: {A}ccelerating facial identification with morphing,'' in
  \emph{International Conference on Acoustics, Speech, and Signal Processing
  ({ICASSP})}.\hskip 1em plus 0.5em minus 0.4em\relax IEEE, May 2019, pp.
  2582--2586.

\bibitem{Drozdowski-SignalFusionWorkload-2021}
P.~Drozdowski, F.~Stockhardt, C.~Rathgeb, and C.~Busch, ``Signal-level fusion
  for indexing and retrieval of facial biometric data,'' \emph{arXiv preprint
  arXiv:2103.03692}, 2021.

\bibitem{Scherhag-MorphingSurvey-2019}
U.~Scherhag, C.~Rathgeb, J.~Merkle, R.~Breithaupt, and C.~Busch, ``Face
  recognition systems under morphing attacks: A survey,'' \emph{IEEE Access},
  vol.~7, pp. 23\,012--23\,026, February 2019.

\bibitem{Wang-indeference-similarity-2017}
Y.~Wang, J.~Wan, J.~Guo, Y.-M. Cheung, and P.~C. Yuen, ``Inference-based
  similarity search in randomized montgomery domains for privacy-preserving
  biometric identification,'' \emph{Transactions on Pattern Analysis and
  Machine Intelligence ({TPAMI})}, vol.~40, no.~7, pp. 1611--1624, July 2017.

\bibitem{OsorioRoig-StableHashFaceIdentification-TBIOM-2021}
D.~Osorio-Roig, C.~Rathgeb, P.~Drozdowski, and C.~Busch, ``Stable hash
  generation for efficient privacy-preserving face identification,''
  \emph{Trans. on Biometrics, Behavior, and Identity Science ({TBIOM})}, 2021.

\bibitem{Patel-CancelableBiometrics-2015}
V.~M. Patel, N.~K. Ratha, and R.~Chellappa, ``Cancelable biometrics: {A}
  review,'' \emph{{IEEE} Signal Processing Magazine}, vol.~32, no.~5, pp.
  54--65, 2015.

\bibitem{BUludag04a}
U.~Uludag, S.~Pankanti, S.~Prabhakar, and A.~K. Jain, ``Biometric
  cryptosystems: issues and challenges,'' \emph{Proceedings of the IEEE},
  vol.~92, no.~6, pp. 948--960, May 2004.

\bibitem{Aguilar-Homomorphic-2013}
C.~Aguilar-Melchor, S.~Fau, C.~Fontaine, G.~Gogniat, and R.~Sirdey, ``Recent
  advances in homomorphic encryption: {A} possible future for signal processing
  in the encrypted domain,'' \emph{{IEEE} Signal Processing Magazine}, vol.~30,
  no.~2, pp. 108--117, 2013.

\bibitem{Nagar10a}
A.~Nagar, K.~Nandakumar, and A.~K. Jain, ``Biometric template transformation: a
  security analysis,'' in \emph{Media Forensics and Security {II}}, vol.
  7541.\hskip 1em plus 0.5em minus 0.4em\relax SPIE, 2010, pp. 237--251.

\bibitem{Wang12a}
Y.~Wang, S.~Rane, S.~C. Draper, and P.~Ishwar, ``A theoretical analysis of
  authentication, privacy, and reusability across secure biometric systems,''
  \emph{Transactions on Information Forensics and Security ({TIFS})}, vol.~7,
  no.~6, pp. 1825--1840, 2012.

\bibitem{Gomez-Barrero-FrameworkForUnlinkability-TemplateProtection-2018}
M.~Gomez-Barrero, J.~Galbally, C.~Rathgeb, and C.~Busch, ``General framework to
  evaluate unlinkability in biometric template protection systems,''
  \emph{Transactions on Information Forensics and Security ({TIFS})}, vol.~13,
  no.~6, pp. 1406--1420, 2018.

\bibitem{ISO-IEC-30136}
{ISO/IEC JTC1 SC37 Biometrics}, \emph{{ISO/IEC} 30136:2018. Information
  technology -- Performance testing of biometric template protection},
  International Organization for Standardization, 2018.

\bibitem{BRatha01a}
N.~K. Ratha, J.~H. Connell, and R.~M. Bolle, ``Enhancing security and privacy
  in biometrics-based authentication systems,'' \emph{IBM Systems Journal},
  vol.~40, no.~3, pp. 614--634, 2001.

\bibitem{Savvides04a}
M.~Savvides, B.~V.~K. {Vijaya Kumar}, and P.~K. Khosla, ``Cancelable biometric
  filters for face recognition,'' in \emph{International Conference on Pattern
  Recognition ({ICPR})}, 2004, pp. 922--925.

\bibitem{Teoh06}
A.~B.~J. Teoh, A.~Goh, and D.~C.~L. Ngo, ``Random multispace quantization as an
  analytic mechanism for biohashing of biometric and random identity inputs,''
  \emph{IEEE Transactions on Pattern Analysis and Machine Intelligence},
  vol.~28, no.~12, pp. 1892--1901, 2006.

\bibitem{Boult06}
T.~Boult, ``Robust distance measures for face-recognition supporting revocable
  biometric tokens,'' in \emph{International Conference on Automatic Face and
  Gesture Recognition ({FGR})}, 2006, pp. 560--566.

\bibitem{GomezBarrero2016}
M.~Gomez-Barrero, C.~Rathgeb, J.~Galbally, C.~Busch, and J.~Fierrez,
  ``Unlinkable and irreversible biometric template protection based on {B}loom
  filters,'' \emph{Information Sciences}, vol. 370--371, pp. 18--32, 2016.

\bibitem{Mai21}
G.~Mai, K.~Cao, X.~Lan, and P.~C. Yuen, ``{SecureFace}: Face template
  protection,'' \emph{Transactions on Information Forensics and Security
  ({TIFS})}, vol.~16, pp. 262--277, 2021.

\bibitem{KONG20061359}
A.~Kong, K.-H. Cheung, D.~Zhang, M.~Kamel, and J.~You, ``An analysis of
  {BioHashing} and its variants,'' \emph{Pattern Recognition}, vol.~39, no.~7,
  pp. 1359--1368, 2006.

\bibitem{BRINGER2017239}
J.~Bringer, C.~Morel, and C.~Rathgeb, ``Security analysis and improvement of
  some biometric protected templates based on {B}loom filters,'' \emph{Image
  and Vision Computing}, vol.~58, pp. 239--253, 2017.

\bibitem{Ghammam20}
L.~Ghammam, K.~Karabina, P.~Lacharme, and K.~{Thiry-Atighehchi}, ``A
  cryptanalysis of two cancelable biometric schemes based on index-of-max
  hashing,'' \emph{Transactions on Information Forensics and Security
  ({TIFS})}, vol.~15, pp. 2869--2880, 2020.

\bibitem{Kirchgasser20b}
S.~Kirchgasser, Y.~Martinez-Diaz, H.~Mendez-Vazquez, and A.~Uhl, ``Is
  warping-based cancellable biometrics (still) sensible for face recognition?''
  in \emph{International Joint Conference on Biometrics ({IJCB})}, 2020, pp.
  1--8.

\bibitem{BJuels99a}
A.~Juels and M.~Wattenberg, ``A fuzzy commitment scheme,'' in \emph{Conference
  on Computer and Communications Security ({CCS})}.\hskip 1em plus 0.5em minus
  0.4em\relax ACM, 1999, pp. 28--36.

\bibitem{BJuels02a}
A.~Juels and M.~Sudan, ``A fuzzy vault scheme,'' in \emph{International
  Symposium on Information Theory ({ISIT})}.\hskip 1em plus 0.5em minus
  0.4em\relax IEEE, 2002, p. 408.

\bibitem{vanderVeen06}
M.~{van der Veen}, T.~Kevenaar, G.-J. Schrijen, T.~H. Akkermans, and F.~Zuo,
  ``Face biometrics with renewable templates,'' in \emph{Security,
  Steganography, and Watermarking of Multimedia Contents {VIII}}, vol.
  6072.\hskip 1em plus 0.5em minus 0.4em\relax SPIE, 2006, pp. 205--216.

\bibitem{Ao09}
M.~Ao and S.~Z. Li, ``Near infrared face based biometric key binding,'' in
  \emph{International Conference on Biometrics ({ICB})}, 2009, pp. 376--385.

\bibitem{Frassen08}
T.~Frassen, X.~Zhou, and C.~Busch, ``Fuzzy vault for 3d face recognition
  systems,'' in \emph{International Conference on Intelligent Information
  Hiding and Multimedia Signal Processing ({IIH-MSP})}, 2008, pp. 1069--1074.

\bibitem{Wang07}
Y.~{Wang} and K.~N. Plataniotis, ``Fuzzy vault for face based cryptographic key
  generation,'' in \emph{Biometrics Symposium}, 2007, pp. 1--6.

\bibitem{rathgeb2021deep}
C.~Rathgeb, J.~Merkle, J.~Scholz, B.~Tams, and V.~Nesterowicz, ``Deep face
  fuzzy vault: Implementation and performance,'' \emph{arXiv preprint
  arXiv:2102.02458}, 2021.

\bibitem{OPM-DataHackNews-2015}
{United States Office of Personnel Management}, ``Statement by {OPM Press
  Secretary Sam Schumach} on background investigations incident,''
  \url{https://www.opm.gov/news/releases/2015/09/cyber-statement-923/},
  September 2015.

\bibitem{Li-ProtectedIndexing-IEEE-2016}
G.~Li, B.~Yang, and C.~Busch, ``A fingerprint indexing algorithm on encrypted
  domain,'' in \emph{Trustcom/BigDataSE/ISPA}.\hskip 1em plus 0.5em minus
  0.4em\relax IEEE, August 2016, pp. 1030--1037.

\bibitem{Drozdowski-ProtectedIrisIndexing-EUSIPCO-2018}
P.~Drozdowski, S.~Garg, C.~Rathgeb, M.~{Gomez-Barrero}, D.~Chang, and C.~Busch,
  ``Privacy-preserving indexing of {Iris-Codes} with cancelable {B}loom
  filter-based search structures,'' in \emph{European Signal Processing
  Conference ({EUSIPCO})}.\hskip 1em plus 0.5em minus 0.4em\relax IEEE,
  September 2018, pp. 1--5.

\bibitem{Murakami-ProtectedIndexing-2019}
T.~Murakami, T.~Ohki, Y.~Kaga, M.~Fujio, and K.~Takahashi, ``Cancelable
  indexing based on low-rank approximation of correlation-invariant random
  filtering for fast and secure biometric identification,'' \emph{Pattern
  Recognition Letters}, vol. 126, pp. 11--20, September 2019.

\bibitem{Pittel-StableRoommates-1994}
P.~G. Pittel and R.~W. Irving, ``An upper bound for the solvability probability
  of a random stable roommates instance,'' \emph{Random Structures \&
  Algorithms}, vol.~5, no.~3, pp. 465--486, July 1994.

\bibitem{Chung-StableRoommates-2000}
K.-S. Chung, ``On the existence of stable roommate matchings,'' \emph{Games and
  economic behavior}, vol.~33, no.~2, pp. 206--230, November 2000.

\bibitem{Rottcher-MorphingDoppelganger-2020}
A.~R{\"o}ttcher, U.~Scherhag, and C.~Busch, ``Finding the suitable
  doppelg{\"a}nger for a face morphing attack,'' in \emph{International Joint
  Conference on Biometrics ({IJCB})}.\hskip 1em plus 0.5em minus 0.4em\relax
  IEEE, September 2020, pp. 1--7.

\bibitem{Kuhn-HungarianAlgorithm-1955}
H.~W. Kuhn, ``The {H}ungarian method for the assignment problem,'' \emph{Naval
  research Logistics Quarterly}, vol.~2, no. 1-2, pp. 83--97, March 1955.

\bibitem{Acar-HESurvey-2018}
A.~Acar, H.~Aksu, A.~S. Uluagac, and M.~Conti, ``A survey on homomorphic
  encryption schemes: Theory and implementation,'' \emph{{ACM} Computer
  Surveys}, vol.~51, no.~4, pp. 79:1--79:35, 2018.

\bibitem{Ricanek-MORPH-2006}
K.~Ricanek and T.~Tesafaye, ``{MORPH}: a longitudinal image database of normal
  adult age-progression,'' in \emph{International Conference on Automatic Face
  and Gesture Recognition ({FGR})}.\hskip 1em plus 0.5em minus 0.4em\relax
  IEEE, April 2006, pp. 341--345.

\bibitem{ICAO-9303-p9-2015}
{International Civil Aviation Organization}, ``Machine readable passports --
  part 9 -- deployment of biometric identification and electronic storage of
  data in {eMRTDs},'' ICAO, Tech. Rep. 9303, 2015.

\bibitem{Deng-ArcFace-2019}
J.~Deng, J.~Guo, and S.~Z.~N. Xue, ``{ArcFace}: Additive angular margin loss
  for deep face recognition,'' in \emph{Conference on Computer Vision and
  Pattern Recognition ({CVPR})}.\hskip 1em plus 0.5em minus 0.4em\relax
  IEEE/CVF, June 2019, pp. 4690--4699.

\bibitem{Huang-CurricularFace-2020}
Y.~Huang, Y.~Wang, Y.~Tai, X.~Liu, P.~Shen, S.~Li, J.~Li, and F.~Huang,
  ``{CurricularFace}: Adaptive curriculum learning loss for deep face
  recognition,'' in \emph{Conference on Computer Vision and Pattern Recognition
  ({CVPR})}.\hskip 1em plus 0.5em minus 0.4em\relax IEEE/CVF, June 2020, pp.
  5901--5910.

\bibitem{Cheon-CKKS-2017}
J.~H. Cheon, A.~Kim, M.~Kim, and Y.~Song, ``Homomorphic encryption for
  arithmetic of approximate numbers,'' in \emph{ASIACRYPT}.\hskip 1em plus
  0.5em minus 0.4em\relax Springer, 2017, pp. 409--437.

\bibitem{DBLP:journals/iacr/FanV12}
J.~Fan and F.~Vercauteren, ``Somewhat practical fully homomorphic encryption,''
  \emph{{IACR} Cryptology ePrint Archive}, vol. 2012, p. 144, 2012.

\bibitem{hoffstein1998ntru}
J.~Hoffstein, J.~Pipher, and J.~H. Silverman, ``{NTRU}: A ring-based public key
  cryptosystem,'' in \emph{International Algorithmic Number Theory
  Symposium}.\hskip 1em plus 0.5em minus 0.4em\relax Springer, 1998, pp.
  267--288.

\bibitem{sealcrypto}
``{M}icrosoft {SEAL} (release 3.6),'' \url{https://github.com/Microsoft/SEAL},
  November 2020, microsoft Research, Redmond, WA.

\bibitem{Kolberg-NTRU-2019}
J.~Kolberg, P.~Bauspie{\ss}, M.~Gomez-Barrero, C.~Rathgeb, M.~D{\"u}rmuth, and
  C.~Busch, ``Template protection based on homomorphic encryption:
  Computationally efficient application to iris-biometric verification and
  identification,'' in \emph{International Workshop on Information Forensics
  and Security ({WIFS})}, December 2019, pp. 1--6.

\bibitem{cryptoeprint:2020:1483}
E.~Crockett, ``A low-depth homomorphic circuit for logistic regression model
  training,'' Cryptology ePrint Archive, Report 2020/1483, 2020,
  \url{https://eprint.iacr.org/2020/1483}.

\bibitem{Zong-HE-2021}
H.~Zong, H.~Huang, and S.~Wang, ``Secure outsourced computation of matrix
  determinant based on fully homomorphic encryption,'' \emph{IEEE Access},
  vol.~9, pp. 22\,651--22\,661, February 2021.

\bibitem{bernstein2017post}
D.~J. Bernstein and T.~Lange, ``Post-quantum cryptography,'' \emph{Nature},
  vol. 549, no. 7671, pp. 188--194, 2017.

\bibitem{gong2019intrinsic}
S.~Gong, V.~N. Boddeti, and A.~K. Jain, ``On the intrinsic dimensionality of
  image representations,'' in \emph{Conference on Computer Vision and Pattern
  Recognition ({CVPR})}.\hskip 1em plus 0.5em minus 0.4em\relax IEEE/CVF, June
  2019, pp. 3987--3996.

\end{thebibliography}

\end{document}